\documentclass[10pt,twocolumn,letterpaper]{article}

\usepackage[pagenumbers]{cvpr} 

\definecolor{cvprblue}{rgb}{0.21,0.49,0.74}
\usepackage[pagebackref,breaklinks,colorlinks,allcolors=cvprblue]{hyperref}

\usepackage{booktabs}
\usepackage{pifont}
\usepackage{adjustbox}
\usepackage{placeins}
\usepackage{svg}
\usepackage{multirow}
\usepackage{graphicx}
\usepackage{tikz}
\usetikzlibrary{arrows.meta,positioning,fit,calc}

\newcommand{\qualrow}[2]{%
    \noindent
    \begin{minipage}[c]{0.025\textwidth}
        \centering
        \rotatebox[origin=c]{90}{\small #1}
    \end{minipage}%
    \begin{minipage}[c]{0.97\textwidth}
        \centering
        \includegraphics[
            width=\linewidth
        ]{#2}
    \end{minipage}%
    \par\vspace{3pt}%
}

\def\paperID{23636} 
\def\confName{CVPR}
\def\confYear{2026}

\title{Body of Evidence: Triangulating 3D Human Pose with \\ Geometry-Conditioned Depth}

\author{
Erik Brorsson$^{1,2}$ \quad
Kilian Freitag$^1$ \quad
Josef Bengtson$^1$ \quad
Kristofer Bengtsson$^2$\\
Lars Hammarstrand$^1$ \quad
Knut \AA kesson$^1$ \quad
Fredrik Kahl$^1$\\[4pt]
$^1$Chalmers University of Technology
\qquad
$^2$Volvo Group\\[3pt]
{\tt\small
\{erik.brorsson,kristofer.bengtsson\}@volvo.com}\\
{\tt\small
\{kilian.freitag,bjosef,lars.hammarstrand,knut.akesson,fredrik.kahl\}@chalmers.se}
}

\begin{document}
\maketitle
\begin{abstract}
Even in multi-camera setups, occlusions can leave individual body joints with only a single usable 2D observation. Such an observation defines a viewing ray but leaves the joint's depth undetermined. We introduce Body of Evidence (BoE), a multi-view 3D human pose estimator that predicts per-camera joint depths alongside 2D joint locations. To reduce monocular depth ambiguity, we propose geometry-conditioned depth prediction, which combines image features with previously reconstructed 3D body geometry as contextual depth cues. We further introduce learnable depth-aware triangulation (LDAT), which augments differentiable algebraic triangulation with predicted depth constraints, enabling reconstruction even from a single contributing camera while preserving the benefits of geometric triangulation when multiple views are available. The reconstructed 3D poses subsequently guide depth prediction in the next iteration, allowing depth estimates and 3D geometry to progressively refine each other. Experiments on CMU Panoptic demonstrate state-of-the-art in-domain performance and superior average performance across evaluated camera subsets and unseen camera arrangements. BoE is particularly effective under sparse camera coverage, outperforming the previous state of the art by 10.6 AP25 points in the challenging three-camera setting.
\end{abstract}    
\section{Introduction}

\begin{figure}[t]
    \centering
    \includegraphics[width=\linewidth]{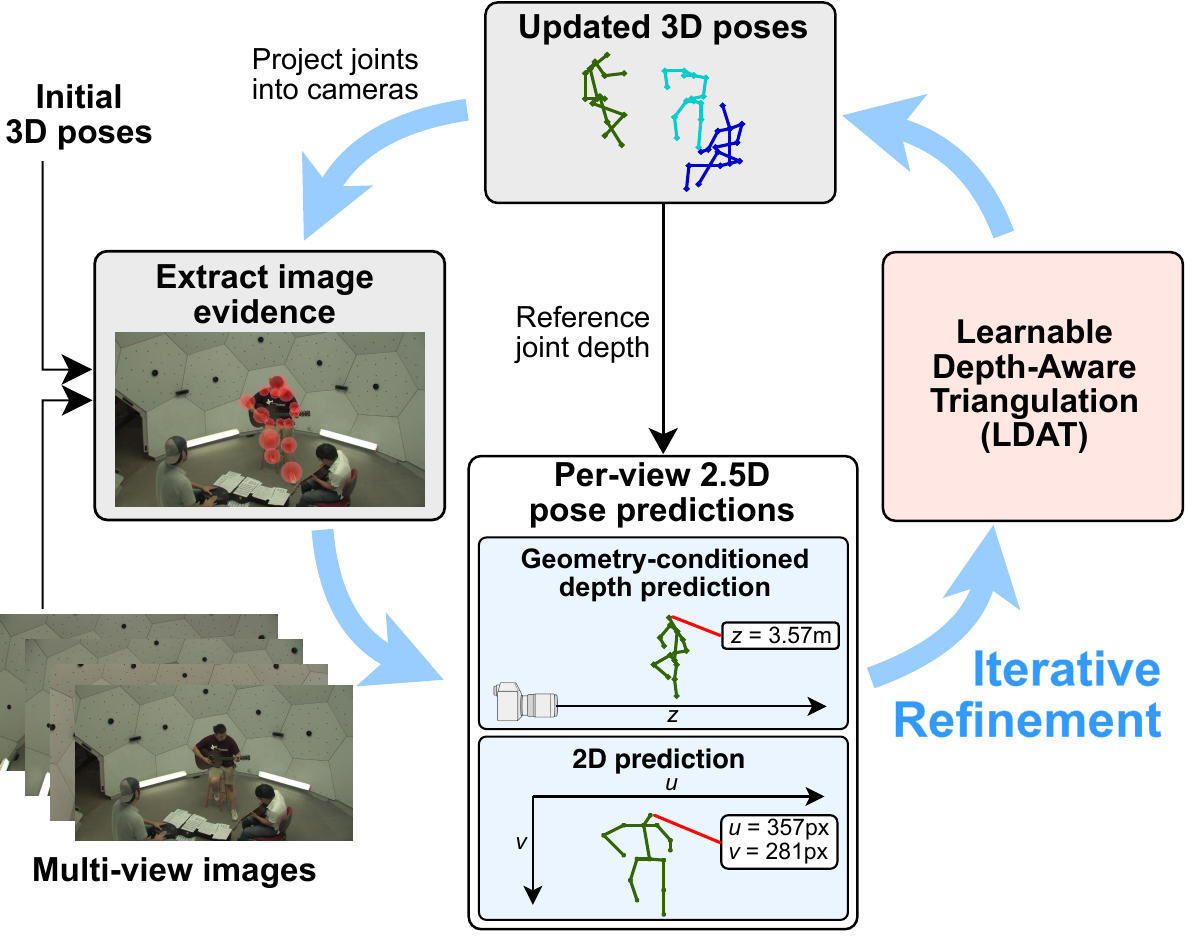}
    \caption{
    \textbf{Overview of BoE.}
Starting from initial 3D pose queries, the decoder
projects joints into calibrated camera views to
extract image evidence.
Each view provides 2D joint and
geometry-conditioned depth predictions, which are
combined through LDAT to update the 3D poses.
The updated poses serve as references for subsequent
decoder layers, enabling iterative refinement.
    }
    \label{fig:overview}
\end{figure}

Estimating 3D human pose from multiple cameras is fundamental to applications such as sports analysis, robotics, surveillance, and industrial monitoring. Traditional approaches
~\cite{bridgeman2019multi,chen2020cross,zhang20204d,dong2019fast,dong2021tracking} decompose the task into monocular 2D pose estimation,
cross-view association, and geometric triangulation.
However, errors in early stages propagate to subsequent
ones, particularly under severe occlusion or sparse
camera coverage, where 2D joint detections may be
unreliable and triangulation becomes poorly conditioned.
End-to-end methods~\cite{mvp,voxelpose} instead learn to predict
3D poses directly from multi-view image features, improving
robustness to occlusions but often exhibiting limited generalization to unseen camera configurations.

Recent methods such as MVGFormer~\cite{mvgformer}
and MV-SSM~\cite{mvssm} combine the advantages
of learned multi-view feature aggregation and
explicit geometry by predicting per-view 2D joint
locations and recovering 3D poses through
differentiable triangulation.
Their iterative refinement framework exploits multi-view
evidence to resolve image-space ambiguities while preserving
the geometric inductive bias of triangulation, supporting
generalization across camera configurations.
Nevertheless, conventional triangulation relies
on observations from multiple views.
When a joint is observed in only one camera,
its 2D location constrains it to a 3D viewing
ray, leaving its position along that ray
undetermined.
Consequently, reconstruction remains challenging
under severe occlusion or limited camera overlap.

We address this limitation with \textbf{Body of Evidence (BoE)},
an end-to-end multi-view 3D human pose estimator
that predicts per-camera 2D joint locations and depths.
We propose \emph{learnable depth-aware triangulation}
(LDAT), which incorporates these predictions as
complementary constraints within a differentiable
algebraic triangulation system.
By combining predicted depths with calibrated
multi-view geometry, LDAT enables 3D reconstruction
even when a joint is observed in only one camera,
while preserving the geometric inductive bias
of triangulation.

A central challenge, however, is the inherent
ambiguity of monocular depth prediction.
While the depth of an individual joint may be
ambiguous from a single view, the estimated 3D
configuration of the body provides metric context
that can help constrain its depth.
To exploit this relationship, we propose
\emph{geometry-conditioned depth prediction},
which conditions individual joint-depth estimates
on previously reconstructed body geometry.
These depth estimates are incorporated into LDAT
to update the 3D poses, which in turn provide
refined geometric context for subsequent depth
predictions.
This iterative feedback allows multi-view geometry
to inform monocular depth estimation, while
learned depth provides additional constraints
for 3D reconstruction.

We summarize our contributions as:
\begin{itemize}

    \item We propose two complementary components: geometry-conditioned depth prediction, which leverages previously reconstructed metric body geometry to reduce monocular depth ambiguity, and LDAT, which incorporates predicted depths into geometric reconstruction, enabling reconstruction even from a single contributing camera.

    \item We introduce BoE, an end-to-end multi-view 3D human pose estimator that iteratively couples depth prediction and triangulation, allowing depth estimates and 3D poses to progressively refine each other.

    \item Experiments on CMU Panoptic demonstrate
    state-of-the-art in-domain performance and
    the highest average AP25 and mAP across
    evaluated camera subsets and unseen camera
    arrangements.
\end{itemize}
\section{Related Work}

\paragraph{Multi-view 3D human pose estimation.}
Classical multi-view methods detect 2D poses independently,
associate them across cameras, and reconstruct 3D joints
through triangulation~\cite{bridgeman2019multi,chen2020cross,zhang20204d,dong2019fast,dong2021tracking}.
While explicit geometry facilitates generalization across
camera configurations, errors in 2D estimation and
cross-view association propagate to the final reconstruction,
particularly under occlusion.
Learnable Triangulation~\cite{iskakov2019learnable}
introduces differentiable algebraic triangulation
into end-to-end learning, enabling 2D predictions to be optimized for 3D reconstruction.
Other methods address reconstruction ambiguities through
3D pictorial structures and graphical models
~\cite{belagiannis2014pictorial,
belagiannis2015revisited,ershadinasab2018multiple},
stochastic triangulation~\cite{bartol2022generalizable},
or explicit anatomical constraints and holistic pose
priors~\cite{chen2022structural,wan2023holistic}.
While structural constraints favor anatomically plausible
reconstructions, they generally cannot uniquely determine
the depth of a joint observed in only one camera.
Alternatively, volumetric methods fuse multi-view
features in a common 3D representation
~\cite{iskakov2019learnable,voxelpose,ye2022faster,zhang2022voxeltrack},
but introduce computational and discretization costs
and may generalize poorly to unseen camera configurations.

Recent transformer-based approaches enable end-to-end
multi-view reasoning without dense 3D volumes.
MvP~\cite{mvp} uses hierarchical queries and attention
to directly estimate 3D poses from multi-view features,
but its learned mapping can depend on the training
camera configuration.
MVGFormer~\cite{mvgformer} instead alternates learned
per-view 2D refinement with explicit geometric triangulation.
MV-SSM~\cite{mvssm} extends this paradigm by modeling
spatial joint dependencies through state-space models.
Both demonstrate strong generalization to unseen camera
configurations, highlighting the benefits of explicit
geometry.
However, since conventional triangulation cannot determine
the depth of a joint observed in only one camera, these methods struggle in sparse camera setups and crowded scenes where occlusions are frequent.
BoE addresses this limitation by predicting
per-view 2.5D observations and incorporating learned
depth constraints into triangulation.

\paragraph{Monocular 3D human pose estimation.}
Monocular 3D pose estimation is inherently ambiguous,
as variations in body scale, articulation, and camera geometry
can produce similar 2D projections.
Many methods focus on root-relative 3D pose estimation,
predicting joint locations relative to a reference joint
~\cite{chen2017matching,martinez2017simple,sun2018integral}.
Alternatively, 2.5D pose representations combine image-plane
joint locations with estimated depths, separating 2D localization
from depth inference~\cite{Iqbal_2018_ECCV,iqbal2020weakly}.
Nie et al.~\cite{nie2017monocular} combine local image evidence
with global skeleton context for joint-depth prediction,
while Jin et al.~\cite{jin2022single} estimate absolute root
and root-relative joint depths using pose-guided depth querying.
Iqbal et al.~\cite{iqbal2020weakly} predict 2.5D poses from
monocular images and exploit multi-view consistency for
weakly supervised learning, without fusing per-view
predictions during 3D reconstruction.
BoE instead estimates per-view 2.5D observations
within an iterative multi-view framework, where previously
triangulated joints provide metric 3D context for depth prediction.

\paragraph{Depth-aware geometric estimation.}
Beyond human pose estimation, learned monocular depth
has been combined with geometric reconstruction
to complement image-based observations.
CNN-SLAM~\cite{tateno2017cnnslam} fuses predicted
depth with geometrically estimated depth, while
DVSO~\cite{yang2018deepvirtual} and
D3VO~\cite{yang2020d3vo} incorporate learned depth
as virtual stereo constraints in photometric
optimization. MP-SfM~\cite{Pataki_2025_CVPR} augments
structure-from-motion with monocular depth and
surface-normal priors, enabling reconstruction of
points observed in a single view and joint refinement
of depth and multi-view scene geometry.
In multi-view 3D human pose estimation,
Luvizon et al.~\cite{luvizon2022consensus} and
Bragagnolo et al.~\cite{bragagnolo2024multi}
independently reconstruct 3D poses from monocular
predictions and subsequently fuse them across views
using per-joint reliability weights.
Luvizon et al. employ predicted joint confidences
and support camera calibration through consensus
optimization, whereas Bragagnolo et al. derive
fusion weights from cross-view reprojection errors
and further refine the fused poses through
reprojection optimization and limb-symmetry constraints.
Lin and Lee~\cite{lin2021planesweep} instead estimate
joint depths from cross-view geometric consistency
and recover 3D poses through back-projection.
In contrast, LDAT jointly incorporates predicted
2D joint locations and absolute camera-frame depths
as complementary constraints within a differentiable
algebraic triangulation system.

To the best of our knowledge, BoE is the first
multi-view 3D human pose estimation framework to
iteratively couple monocular depth prediction with
depth-aware triangulation, enabling bidirectional
refinement through geometry-conditioned depth estimation
and reconstruction.

\section{Methodology}
\label{sec:method}

\subsection{Problem Formulation}

Given $V$ synchronized images $\{I_v\}_{v=1}^{V}$ from calibrated
cameras with intrinsic matrices $K_v$ and extrinsic parameters
$[R_v \mid t_v]$, our goal is to estimate the 3D poses of
all $M$ people in the scene.
Each pose consists of $J$ joints,
$\mathbf{X}_m = \{\mathbf{x}_{m,j}\}_{j=1}^{J}$,
where $\mathbf{x}_{m,j} \in \mathbb{R}^3$ denotes the 3D
location of joint $j$ of person $m$ in a common world
coordinate system.
The number of people $M$ is unknown and must be inferred
from the multi-view observations.

\subsection{Overview}

\begin{figure*}
    \centering
    \includegraphics[width=1\linewidth]{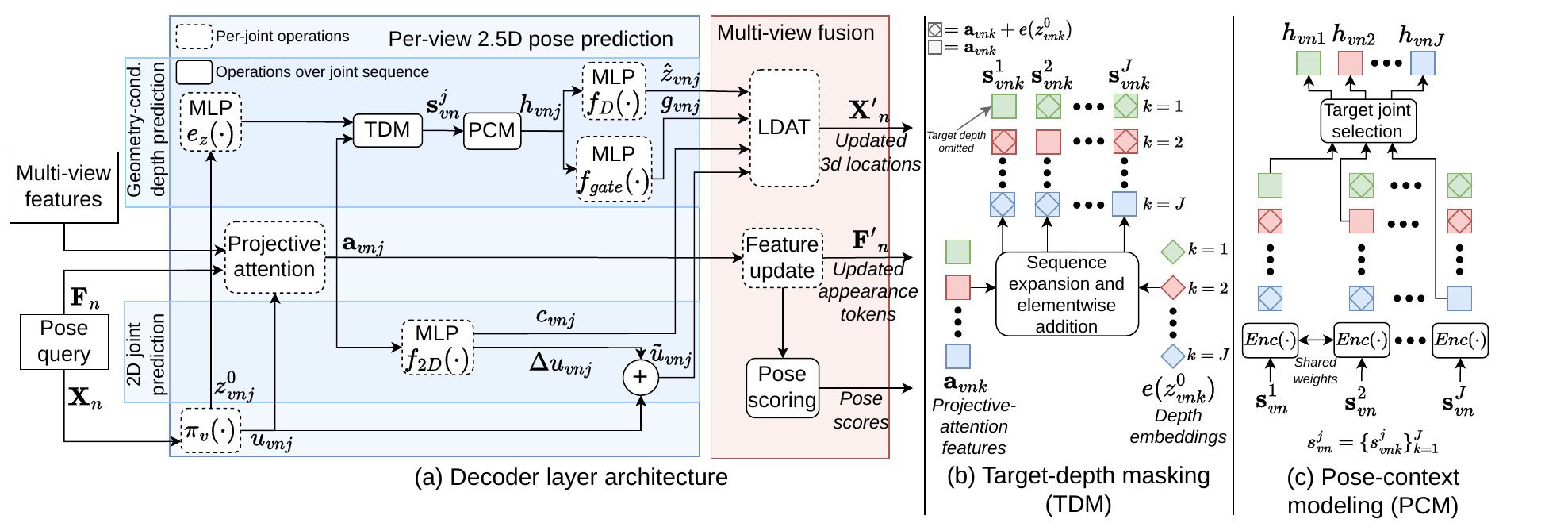}
    \caption{BoE decoder architecture and geometry-conditioned
depth estimation.
(a) Each decoder layer extracts image evidence through
projective attention and predicts per-view 2D joint
locations, confidences, depths, and depth gates.
Learnable depth-aware triangulation (LDAT) fuses these
measurements to update the 3D poses, while separate
branches update appearance features and predict pose scores.
(b) Target-depth masking (TDM) constructs a separate
joint sequence for each target joint, excluding only
its own reference-depth embedding while retaining
its image feature.
(c) Pose-context modeling (PCM) processes the masked
sequences using a shared encoder and selects the
contextualized feature of each target joint for
depth prediction.}
    \label{fig:method}
\end{figure*}

Our architecture follows an iterative multi-view pose
estimation framework~\cite{mvgformer,mvssm}, in which
3D pose queries are progressively refined through
per-view prediction and multi-view fusion. The decoder maintains $N$ pose queries, each representing
a candidate person through $J$ 3D joint locations
$\mathbf{X}_n = \{\mathbf{x}_{n,j}\}_{j=1}^{J}$
and corresponding $D$-dimensional appearance tokens
$\mathbf{F}_n = \{\mathbf{f}_{n,j}\}_{j=1}^{J}$,
where $\mathbf{f}_{n,j}\in\mathbb{R}^{D}$.
The appearance tokens are initialized using learned
person and joint embeddings~\cite{mvp,mvgformer,mvssm},
while the initial 3D poses are constructed by translating
a fixed T-pose template to root locations sampled
on a regular 2D grid.

As illustrated in Fig.~\ref{fig:method}(a),
each decoder layer consists of two stages.
First, per-view 2.5D pose prediction extracts image
features through projective attention and predicts
2D joint locations, camera-frame depths, and their
confidence weights.
Second, multi-view fusion combines these predictions
through LDAT
to update the 3D poses.
The updated geometry and appearance features are
passed to the next decoder layer for further refinement.

\subsection{Per-view 2.5D Pose Prediction}
\label{sec:2.5d-prediction}

At each decoder layer, every 3D joint $\mathbf{x}_{n,j}$
is transformed into the coordinate system of camera $v$,
yielding its camera-frame reference depth $z^0_{v,n,j}$,
and subsequently projected onto the image plane to obtain
its image coordinate $\mathbf{u}_{v,n,j}$.
For brevity, we jointly denote these operations by $\pi_v$:
\begin{equation}
\begin{aligned}
(\mathbf{u}_{v,n,j},z^0_{v,n,j})
&= \pi_v(\mathbf{x}_{n,j}) \\
&= \bigl(
\operatorname{proj}(K_v\mathbf{q}_{v,n,j}),
[\mathbf{q}_{v,n,j}]_3
\bigr),
\end{aligned}
\end{equation}
where $\mathbf{q}_{v,n,j}=R_v\mathbf{x}_{n,j}+t_v$
is the joint in camera coordinates, $[\cdot]_3$ denotes
the third vector component, and
$\operatorname{proj}([x,y,z]^\top)=(x/z,y/z)^\top$
denotes perspective division.
Projective attention~\cite{mvp} samples multi-scale image features around
$\mathbf{u}_{v,n,j}$ and produces a view-specific feature
$\mathbf{a}_{v,n,j}\in\mathbb{R}^{D}$.

\paragraph{2D joint prediction.}
A shared three-layer MLP with ReLU activations
predicts a 2D correction and a confidence score from the
projective-attention feature:
\begin{equation}
[\Delta\mathbf{u}_{v,n,j}, c_{v,n,j}]
=
f_{\mathrm{2D}}(\mathbf{a}_{v,n,j}).
\label{eq:2d-head}
\end{equation}
The predicted correction is added to the projected
reference coordinate:
\begin{equation}
\tilde{\mathbf{u}}_{v,n,j}
=
\mathbf{u}_{v,n,j}
+
\Delta\mathbf{u}_{v,n,j}.
\end{equation}
The confidence scores $c_{v,n,j}$ are softmax-normalized
across valid camera views.

\paragraph{Geometry-conditioned depth prediction.}
Monocular depth estimation is inherently ambiguous, as different
3D configurations can produce similar image observations.
However, at later decoder layers, joints reliably observed
by multiple cameras can provide accurate metric depth references
for joints with weaker observations.
We therefore condition depth prediction on the appearance
and estimated geometry of the remaining joints of the same person, as detailed below.

\noindent\emph{Depth embedding (DE).}
For each joint, a two-layer MLP $e_z(\cdot)$ maps its
camera-frame reference depth $z^0_{v,n,j}$, expressed
in metres, to a $D$-dimensional embedding.
Adding this embedding to the projective-attention feature
provides the depth branch with geometric information
obtained from previous decoder layers.

\noindent\emph{Target-Depth Masking (TDM).}
Directly providing the target joint's reference depth
could encourage the network to copy its current estimate
rather than infer depth from the image.
To prevent this shortcut, we construct a separate
joint sequence for each target joint $j$, as illustrated
in Fig.~\ref{fig:method}(b):
\begin{equation}
\mathbf{s}^{(j)}_{v,n,k}
=
\begin{cases}
\mathbf{a}_{v,n,k}+e_z(z^0_{v,n,k}),
& k\neq j,\\
\mathbf{a}_{v,n,j},
& k=j.
\end{cases}
\label{eq:masked-depth-embedding}
\end{equation}
Thus, the target joint retains its image feature but not
its own reference-depth embedding, while all other joints
provide appearance and geometric context.

\noindent\emph{Pose-context modeling (PCM).}
As illustrated in Fig.~\ref{fig:method}(c),
each masked sequence is processed by a joint-feature
encoder consisting of multi-head self-attention
and feed-forward updates.
The contextualized feature for target joint $j$ is
\begin{equation}
\mathbf{h}_{v,n,j}
=
\left[
\operatorname{Enc}
\left(
\{\mathbf{s}^{(j)}_{v,n,k}\}_{k=1}^{J}
\right)
\right]_j.
\label{eq:depth-contextualized-feature}
\end{equation}
The encoder operates independently for each person query
and camera, allowing interactions among the joints of one
skeleton without mixing information across people or views.

\noindent\emph{Depth and confidence prediction.}
Two independent three-layer MLPs predict the absolute
camera-frame depth and its confidence gate from the
contextualized feature $\mathbf{h}_{v,n,j}$:
\begin{equation}
\begin{aligned}
\hat z_{v,n,j} &= 1000\,f_{\mathrm{D}}(\mathbf{h}_{v,n,j}),
\qquad
g_{v,n,j} &= \sigma\left(f_{\mathrm{gate}}(\mathbf{h}_{v,n,j})\right).
\end{aligned}
\label{eq:depth-prediction}
\end{equation}
The depth prediction is expressed in metres and converted
to millimetres for geometric computation.
Unlike the 2D confidences, which are softmax-normalized
across cameras, the depth gates are independent sigmoid
outputs. This allows multiple depth measurements to contribute
simultaneously, while also permitting unreliable depth
estimates to be suppressed.

Together, the predicted 2D location and depth form a
per-view 2.5D joint measurement
$(\tilde{\mathbf{u}}_{v,n,j},\hat z_{v,n,j})$,
with separate confidence weights for the image-space
and depth constraints.

\subsection{Multi-view Fusion}
\label{sec:multiview-fusion}

\paragraph{Learnable depth-aware triangulation (LDAT).}
Conventional triangulation estimates each 3D joint using only
its multi-view 2D observations. Consequently, its geometric
constraints are insufficient when a joint is observed in
only one camera. We address this limitation by incorporating the predicted
camera-frame depths directly into the triangulation system. Thus, even a single view can constrain the 3D joint
location when a reliable depth estimate is available.
For clarity, we omit person and joint indices below.

Let $P_v=K_v[R_v\mid t_v]$ denote the projection matrix of
camera $v$, and let $\mathbf{p}_{v,k}^{\top}$ be its $k$-th row.
For a homogeneous 3D point
$\tilde{\mathbf{x}}=(X,Y,Z,1)^{\top}$,
the corrected 2D observation
$\tilde{\mathbf{u}}_v=(\tilde u_v,\tilde v_v)$
provides the standard DLT constraints:
\begin{equation}
\begin{aligned}
(\tilde u_v\mathbf{p}_{v,3}^{\top}
-\mathbf{p}_{v,1}^{\top})\tilde{\mathbf{x}} &= 0,\\
(\tilde v_v\mathbf{p}_{v,3}^{\top}
-\mathbf{p}_{v,2}^{\top})\tilde{\mathbf{x}} &= 0.
\end{aligned}
\label{eq:dlt-rows}
\end{equation}
Both constraints are weighted by the predicted
2D confidence $c_v$.
Our depth-aware triangulation introduces an additional
constraint for each view, weighted by $g_v$:
\begin{equation}
\left[
\mathbf{r}_{v,3}^{\top},
\;t_{v,3}-\hat z_v
\right]\tilde{\mathbf{x}}
=0,
\label{eq:depth-row}
\end{equation}
where $\mathbf{r}_{v,3}^{\top}$ and $t_{v,3}$
are the third row of $R_v$ and the third component of $t_v$,
respectively.

We stack the weighted 2D and depth constraints from all
valid views into a matrix $A$ and solve:
\begin{equation}
\tilde{\mathbf{x}}^*
=
\underset{\|\tilde{\mathbf{x}}\|_2=1}{\arg\min}
\|A\tilde{\mathbf{x}}\|_2^2.
\label{eq:depth-aware-dlt}
\end{equation}
The solution is obtained from the right singular vector
of $A$ associated with its smallest singular value,
followed by homogeneous normalization.

\paragraph{Pose query update.}
We aggregate the projective-attention
features across camera slots and fuse them with the existing appearance token $\mathbf{f}_{n,j}$ in a feed-forward block~\cite{mvgformer}:
\begin{equation}
\bar{\mathbf{a}}_{n,j}
=
\frac{1}{V}
\sum_{v=1}^{V}
\mathbf{a}_{v,n,j},
\qquad
\mathbf{f}'_{n,j}
=
\operatorname{FFN}
(\mathbf{f}_{n,j},\bar{\mathbf{a}}_{n,j}).
\label{eq:query-feature-update}
\end{equation}
The update uses linear projection and feed-forward layers
with residual connections and layer normalization.
The updated appearance token and triangulated 3D joint
are passed to the next decoder layer.

\paragraph{Pose scoring (PS).}
To assess the quality of each reconstructed pose,
we predict a confidence score from the joint
appearance tokens of the complete pose query. Specifically, we prepend a learned classification token to the sequence
of $J$ joint appearance tokens and process the sequence
using multi-head self-attention.
The classification-token output is mapped to a scalar
pose confidence, allowing the model to assess the
joint evidence of the complete pose.
Low-scoring hypotheses are filtered between decoder
layers and during final post-processing.

\subsection{Training Objective}
\label{sec:training}

We adopt the fixed many-to-one anchor-based assignment
of~\cite{mvgformer} for pose regression, but replace
its anchor-based classification targets with labels
determined by reconstruction accuracy.
Specifically, a predicted pose is labeled positive
if its MPJPE to the closest ground-truth pose falls
below a threshold.
These labels are recomputed at each decoder layer,
encouraging pose scores to reflect reconstruction
accuracy rather than initial anchor proximity.

At each decoder layer, we supervise the
regression-matched queries using 3D pose, 2D
joint-location, and depth losses, while the
classification loss supervises all queries using
their layer-specific labels.
The per-layer training objective is
\begin{equation}
\mathcal{L}
=
\lambda_{\mathrm{cls}}\mathcal{L}_{\mathrm{cls}}
+
\lambda_{\mathrm{3D}}\mathcal{L}_{\mathrm{3D}}
+
\lambda_{\mathrm{2D}}\mathcal{L}_{\mathrm{2D}}
+
\lambda_{\mathrm{D}}\mathcal{L}_{\mathrm{depth}}.
\label{eq:objective}
\end{equation}
The depth loss measures the $\ell_1$ error between
predicted and ground-truth absolute camera-frame
joint depths, excluding joints whose ground-truth
projections fall outside the image bounds.
All decoder layers are supervised; further loss
details are provided in the supplementary material.

\paragraph{DropView augmentation (DV).}
To improve robustness under sparse-view conditions,
we randomly reduce the number of available cameras
during training.
For each mini-batch, we uniformly sample the number
of retained cameras between a configured minimum
and the total number of available cameras.
This exposes the model to varying observation conditions
and encourages it to combine multi-view geometry with
monocular depth cues when only limited camera evidence
is available.
\newcommand{\cmark}{\ding{51}}
\newcommand{\xmark}{\ding{55}}

\section{Experiments}

\subsection{Experimental Setup}

\paragraph{Datasets.}
We follow the evaluation protocol of~\cite{mvgformer,mvssm},
training exclusively on CMU Panoptic~\cite{joo2015panoptic}
with the standard five-camera CMU0 configuration and train--test split.
To evaluate generalization to different numbers of cameras,
we denote by CMU0($K$) the CMU0 configuration evaluated
with $K$ cameras, removing or adding cameras at inference
relative to the training configuration.
For cross-arrangement evaluation, we test on
CMU1--CMU4, which use camera arrangements unseen
during training.
We additionally evaluate cross-scene generalization
on Campus~\cite{belagiannis2014pictorial} and Shelf~\cite{belagiannis2014pictorial}.
We use CMU3 alongside CMU0 for ablation studies
and CMU3 for qualitative comparisons, as it has
the fewest cameras among CMU1--CMU4 and provides
a challenging setting for evaluating unseen camera
arrangements under limited camera coverage.

\paragraph{Evaluation metrics.}
On CMU Panoptic, we report Average Precision at a 25mm threshold (AP$25$), mean Average Precision (mAP), and Mean Per Joint Position Error (MPJPE), following prior work~\cite{mvgformer,mvssm}. mAP is computed by averaging AP over MPJPE thresholds of ${25,50,75,100,125,150}$. MPJPE is reported in millimeters. For Campus and Shelf, we report Percentage of Correct Parts (PCP) following the standard evaluation protocol.

\paragraph{Implementation details.}
Following prior works~\cite{voxelpose,mvp,mvgformer,mvssm},
we use a ResNet-50~\cite{he2016deep} backbone pretrained
for 2D pose estimation~\cite{xiao2018simple} on COCO.
We use four decoder layers and follow~\cite{mvgformer}
for query initialization, hidden dimension, confidence
filtering, and non-maximum suppression.
We train for 40 epochs with a learning rate of $4\times10^{-4}$
and early stopping.
To stabilize training, we employ a four-epoch warmup
during which the depth branch is supervised, but its
predictions are excluded from LDAT and DropView is disabled.
After warmup, LDAT and DropView are
enabled, with DropView retaining at least one camera.
All baseline results in the comparisons with prior work
are taken from MVGFormer~\cite{mvgformer} and
MV-SSM~\cite{mvssm}.
Results for the full model are averaged over three independent trainings with different random seeds,
while ablations use a single run and seed.
Further implementation details are provided in Sec.~\ref{sec:suppl_arch}
of the supplementary material.

\subsection{In-domain results}
Table~\ref{tab:paper-results-1} compares our method with
prior work on the standard five-camera CMU Panoptic
configuration, CMU0(5), using the conventional
train--test split. BoE achieves the highest AP25 (93.7) and
lowest MPJPE (15.4~mm), improving over MVGFormer by 1.4 AP25 points and 0.6~mm,
respectively. Compared with MV-SSM, BoE improves AP25 by 0.2 points
and reduces MPJPE by 0.3~mm.
These results indicate that incorporating monocular depth cues into
geometric reconstruction can improve pose accuracy even under the
five-camera training configuration.

\begin{table}[t]
  \centering
  \caption{In-domain results on CMU Panoptic with conventional train-test split. A dash indicates information not reported by the model.}
  \label{tab:paper-results-1}
  \small
  \setlength{\tabcolsep}{3pt}
  \begin{adjustbox}{max width=\columnwidth}
  \begin{tabular}{llll}
    \toprule
    Method & Venue & AP25 $\uparrow$ & MPJPE $\downarrow$ \\
    \midrule
    VoxelPose \cite{voxelpose} & ECCV'20 & 84.0 & 17.7 \\
    Lin et al. \cite{lin2021planesweep} & CVPR'21 & 92.1 & 16.8 \\
    Graph3D \cite{wu2021graph} & ICCV'21 & -- & 15.8 \\
    MvP \cite{mvp} & NeurIPS'21 & 92.3 & 15.8 \\
    Faster VoxelPose \cite{ye2022faster} & ECCV'22 & 85.2 & 18.3 \\
    MVGFormer \cite{mvgformer} & CVPR'24 & 92.3 & 16.0 \\
    MV-SSM \cite{mvssm} & CVPR'25 & 93.5 & 15.7 \\
    BoE & Ours & $\mathbf{93.7} \pm 0.4$ & $\mathbf{15.4} \pm 0.2$ \\
    \bottomrule
  \end{tabular}
  \end{adjustbox}
\end{table}

\subsection{Cross-camera and cross-arrangement generalization}

\begin{table*}[t]
  \centering
  \caption{Cross-camera and cross-arrangement generalization.
Models are trained on CMU0 with five cameras and evaluated
without fine-tuning. Cross-camera averages are computed over
CMU0($K$), $K=3$ to $7$, including the in-domain configuration
CMU0(5), whose AP25 results are reported in
Table~\ref{tab:paper-results-1}.
Cross-arrangement averages are computed over CMU1--CMU4.}
  \label{tab:paper-results-3}
  \label{tab:paper-results-4}
  \small
  \setlength{\tabcolsep}{3pt}
  \begin{adjustbox}{max width=\textwidth}
  \begin{tabular}{cl|ll|ll|ll|ll|ll}
    \toprule
    & Method & \multicolumn{2}{c|}{CMU0 (3)} & \multicolumn{2}{c|}{CMU0 (4)} & \multicolumn{2}{c|}{CMU0 (6)} & \multicolumn{2}{c|}{CMU0 (7)} & \multicolumn{2}{c}{Average} \\
    & & AP25 $\uparrow$ & mAP $\uparrow$ & AP25 $\uparrow$ & mAP $\uparrow$ & AP25 $\uparrow$ & mAP $\uparrow$ & AP25 $\uparrow$ & mAP $\uparrow$ & AP25 $\uparrow$ & mAP $\uparrow$ \\
    \midrule
    \multirow{6}{*}{\rotatebox{90}{Cross-camera}} & Dong et al. \cite{dong2019fast} & 0.0 & 37.7 & 0.0 & 46.6 & 0.1 & 41.8 & 0.2 & 40.1 & 13.3 & 41.6 \\
     & VoxelPose \cite{voxelpose} & 13.7 & 73.0 & 65.4 & 93.1 & 84.5 & 96.8 & 81.5 & 96.3 & 69.1 & 89.8 \\
     & MvP \cite{mvp} & 12.3 & 57.1 & 59.8 & 84.1 & 0.0 & 0.3 & 0.0 & 0.2 & 21.6 & 35.4 \\
     & MVGFormer \cite{mvgformer} & 44.6 & 83.4 & 83.6 & 96.0 & 94.7 & 98.5 & 95.1 & 98.6 & 83.3 & 94.1 \\
     & MV-SSM \cite{mvssm} & 55.4 & 90.3 & 84.9 & 97.0 & 94.6 & \textbf{98.8} & \textbf{95.6} & \textbf{98.8} & 84.8 & 96.7 \\
     & BoE & $\mathbf{66.0} \pm 1.8$ & $\mathbf{92.9} \pm 0.3$ & $\mathbf{87.7} \pm 0.5$ & $\mathbf{97.5} \pm 0.1$ & $\mathbf{95.1} \pm 0.4$ & $\mathbf{98.8} \pm 0.1$ & $95.2 \pm 0.4$ & $98.7 \pm 0.2$ & $\mathbf{87.5} \pm 0.4$ & $\mathbf{97.3} \pm 0.1$ \\
    \midrule
    & Method & \multicolumn{2}{c|}{CMU1} & \multicolumn{2}{c|}{CMU2} & \multicolumn{2}{c|}{CMU3} & \multicolumn{2}{c|}{CMU4} & \multicolumn{2}{c}{Average} \\
    \midrule
    \multirow{6}{*}{\rotatebox{90}{Cross-arrange}} & Dong et al. \cite{dong2019fast} & 0.1 & 39.9 & 0.0 & 29.2 & 0.0 & 29.2 & 0.0 & 32.8 & 0.0 & 32.8 \\
     & VoxelPose \cite{voxelpose} & 28.8 & 86.9 & 26.9 & 69.6 & 21.0 & 69.6 & 40.9 & 87.9 & 29.4 & 78.5 \\
     & MvP \cite{mvp} & 0.0 & 0.0 & 0.0 & 0.0 & 0.0 & 0.0 & 0.0 & 0.0 & 0.0 & 0.0 \\
     & MVGFormer \cite{mvgformer} & 86.8 & 96.1 & 67.9 & 90.3 & 52.5 & 78.8 & 91.5 & 97.2 & 74.7 & 90.6 \\
     & MV-SSM \cite{mvssm} & 86.0 & 97.0 & 74.6 & 93.5 & 59.5 & 84.8 & 94.2 & \textbf{98.5} & 78.6 & 93.5 \\
     & BoE & $\mathbf{89.4} \pm 0.2$ & $\mathbf{97.0} \pm 0.1$ & $\mathbf{82.0} \pm 0.5$ & $\mathbf{95.7} \pm 0.1$ & $\mathbf{62.0} \pm 4.3$ & $\mathbf{91.6} \pm 0.9$ & $\mathbf{94.6} \pm 0.3$ & $98.4 \pm 0.2$ & $\mathbf{82.0} \pm 1.1$ & $\mathbf{95.7} \pm 0.2$ \\
    \bottomrule
  \end{tabular}
  \end{adjustbox}
\end{table*}

We first evaluate generalization to different
numbers of cameras using CMU0($K$). Following prior work~\cite{mvgformer,mvssm},
we compute averages over $K=3$ to $7$, including
the in-domain configuration CMU0(5).
As shown in Table~\ref{tab:paper-results-3},
BoE achieves the highest average
AP25 (87.5) and mAP (97.3), outperforming MV-SSM
by 2.7 and 0.6 points, respectively.
The largest improvements occur with fewer cameras.
On CMU0(3), BoE achieves 66.0 AP25
and 92.9 mAP, exceeding MV-SSM by 10.6
and 2.6 points.
On CMU0(4), the corresponding improvements
are 2.8 and 0.5 points.
With six or seven cameras, performance remains
competitive with the strongest baselines.
These results suggest that the additional
monocular depth constraints are particularly
beneficial when multi-view geometric
observations are limited.
This is further supported by the analysis in
Sec.~\ref{sec:suppl_da_triangulation} of the supplementary
material, which shows that LDAT
provides the largest benefits for joints with fewer
contributing cameras.

We further evaluate generalization to unseen
camera arrangements using CMU1--CMU4.
As shown in Table~\ref{tab:paper-results-4},
BoE achieves the highest average
AP25 (82.0) and mAP (95.7),
surpassing MV-SSM by 3.4 and 2.2
points, respectively.
The largest AP25 improvement occurs
on CMU2, where BoE reaches 82.0
compared with 74.6 for MV-SSM.
On CMU3, BoE improves mAP from
84.8 to 91.6, while AP25 increases
from 59.5 to 62.0.
BoE also achieves the highest
AP25 on CMU1 and CMU4, although
MV-SSM retains a slightly higher mAP
on CMU4.
Overall, these results indicate
strong generalization to camera
configurations not observed during training.
Qualitative comparisons on CMU3 are shown in
Figs.~\ref{fig:qual-rgb} and~\ref{fig:qual-3d},
with additional visualizations provided in
the supplementary material.

\begin{figure*}
    \centering
    \qualrow{\hspace{15pt} Ours \hspace{32pt} MVGFormer}{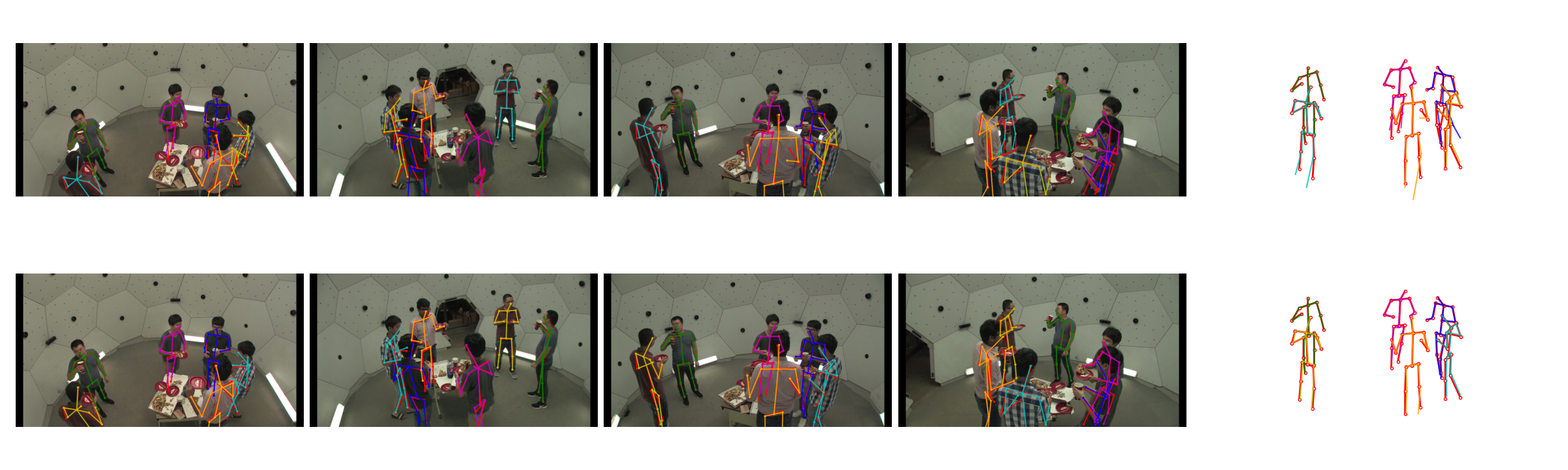}
\caption{\textbf{Qualitative comparison of BoE and
MVGFormer~\cite{mvgformer} on CMU3.}
RGB images with predicted 2D poses and projected
ground-truth poses are shown on the left, and
reconstructed 3D poses on the right.
Ground truth is shown in red, while predictions
are shown in other colors.
BoE more accurately reconstructs heavily occluded
joints, including the right hand and foot of the
orange pose, the feet of the turquoise pose, and
the left leg of the blue pose (colors refer to
the top-row predictions).}
    \label{fig:qual-rgb}
\end{figure*}

\begin{figure}
    \centering
    \includegraphics[width=1.0\linewidth]{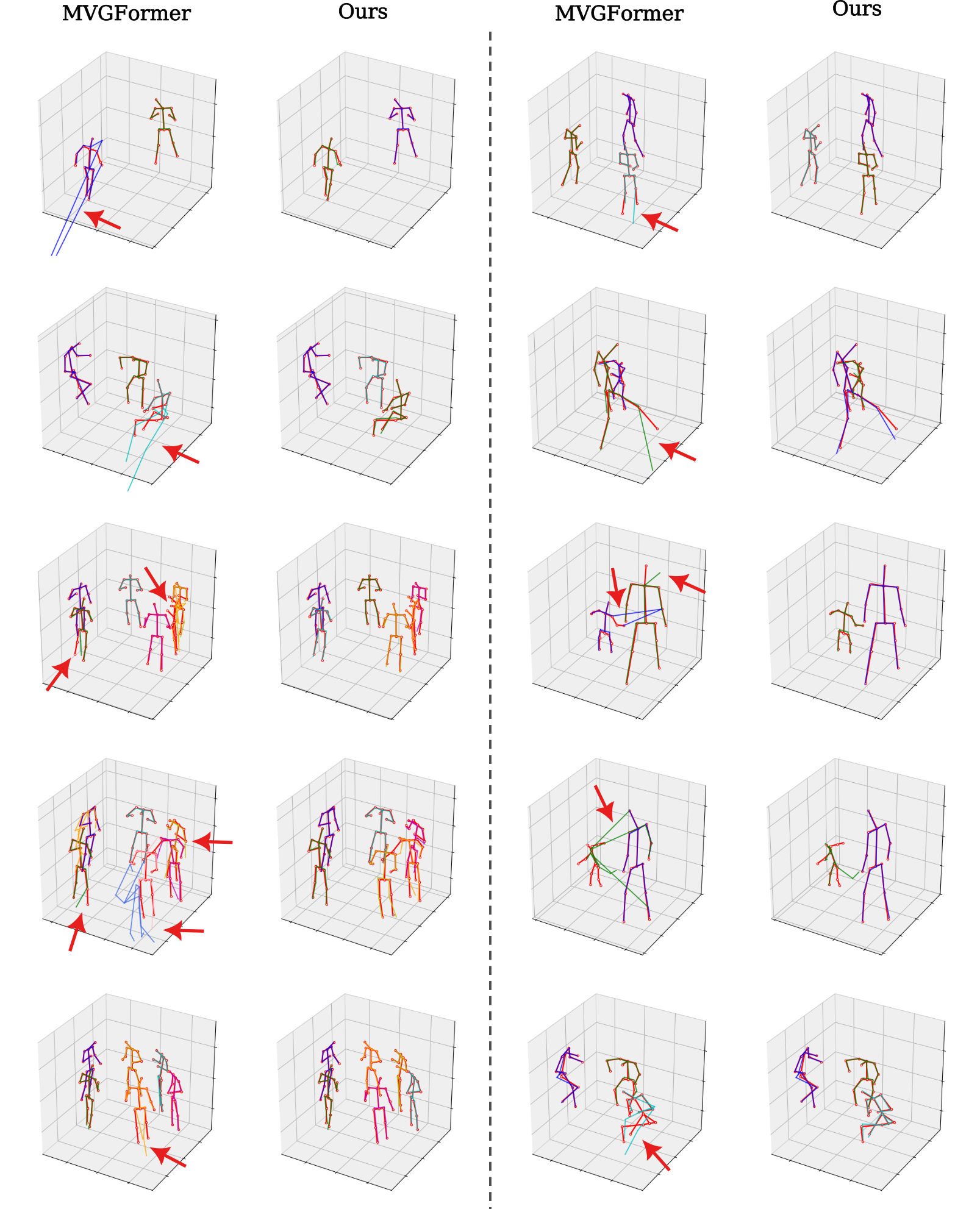}
\caption{\textbf{Qualitative comparison of BoE and
MVGFormer~\cite{mvgformer} on CMU3.}
Ten validation examples are shown, with ground-truth
poses in red and predictions in other colors.
BoE generally reconstructs more accurate 3D poses,
particularly for joints with limited multi-view support.
Corresponding RGB images and predicted 2D poses are
provided in supplementary Sec.~\ref{sec:suppl_qualitative}.}
    \label{fig:qual-3d}
\end{figure}

\subsection{Cross-scene generalization}
Table~\ref{tab:paper-results-2} evaluates
cross-scene transfer from CMU Panoptic to Campus
and Shelf.
On Shelf, BoE achieves 87.7 average PCP,
comparable to MVGFormer (87.9) and MV-SSM (88.0).
However, performance drops substantially on Campus,
where BoE achieves 26.6 average PCP,
compared with 58.1 for MVGFormer and 67.3 for MV-SSM.
Further analysis in Sec.~\ref{sec:suppl_depth_analysis}
of the supplementary material reveals a substantial
increase in mean depth-prediction error, from
27.5~mm on CMU0 and 34.1~mm on CMU3
to 61.7~mm on Shelf and 6,583.4~mm on Campus.
This coincides with a shift in typical subject
depths from 2--5~m on CMU Panoptic to 5--15~m
on Campus, suggesting that the learned depth
predictions are sensitive to substantial
depth-distribution shifts.

The modular pipeline of~\cite{dong2019fast}
outperforms all evaluated end-to-end methods \cite{mvp,voxelpose,mvgformer,mvssm} on both
datasets.
These results highlight the distinction between
generalization to unseen camera configurations
and transfer across scenes, where changes in
appearance and environmental conditions introduce
additional challenges.

\begin{table}[t]
  \centering
  \caption{Cross-scene generalization evaluation. All models were trained on CMU0 and evaluated on Campus and Shelf without fine-tuning.}
  \label{tab:paper-results-2}
  \small
  \setlength{\tabcolsep}{3pt}
  \begin{adjustbox}{max width=\columnwidth}
  \begin{tabular}{l|l|l}
    \toprule
    Method & Campus Avg. PCP $\uparrow$ & Shelf Avg. PCP $\uparrow$ \\
    \midrule
    Dong et al. \cite{dong2019fast} & \textbf{96.3} & \textbf{96.9} \\
    MvP \cite{mvp} & 0.0 & 8.7 \\
    VoxelPose \cite{voxelpose} & 11.2 & 69.8 \\
    MVGFormer \cite{mvgformer} & 58.1 & 87.9 \\
    MV-SSM \cite{mvssm} & 67.3 & 88.0 \\
    BoE & $26.6 \pm 1.2$ & $87.7 \pm 0.0$ \\
    \bottomrule
  \end{tabular}
  \end{adjustbox}
\end{table}

\subsection{Ablations}
Table~\ref{tab:paper-results-5} evaluates
the contributions of the depth branch (DB),
pose scoring (PS), and DropView augmentation
(DV) on CMU0 and CMU3.
Removing DB eliminates monocular depth prediction
and replaces LDAT with conventional 2D triangulation.
When ablating PS, we revert to
the classification strategy of~\cite{mvgformer},
which predicts joint-wise classification scores
and uses fixed anchor-based classification targets.
Removing DV disables camera-dropout augmentation
during training.

The complete model (row~(f)) achieves the highest
AP25 and mAP on both datasets.
Removing DB (row~(c)) reduces AP25/mAP by
1.3/0.8 points on CMU0 and 18.3/9.6 points on CMU3,
demonstrating that depth-aware reconstruction
is particularly beneficial under challenging
camera configurations.
Removing PS (row~(d)) reduces AP25/mAP by
9.9/2.4 points on CMU0 and 17.9/7.2 points on CMU3,
highlighting the importance of pose-level confidence
estimation in both settings.
Similarly, removing DV (row~(e)) reduces AP25/mAP by
2.1/0.5 points on CMU0 and 10.8/4.2 points on CMU3,
indicating that camera-dropout augmentation
is particularly beneficial for cross-arrangement
generalization.

When all three components are disabled (row~(a)),
our framework reduces to MVGFormer.
Adding PS and DV without DB (rows~(b)--(c))
does not consistently improve performance
over this baseline.
In contrast, both components substantially
benefit the complete depth-aware framework,
suggesting that their effectiveness depends
on the underlying reconstruction architecture.

\begin{table}[t]
  \centering
  \caption{Ablations of the main components: pose scoring (PS), DropView (DV), and the depth branch (DB).}
  \label{tab:paper-results-5}
  \small
  \setlength{\tabcolsep}{3pt}
  \begin{adjustbox}{max width=\columnwidth}
  \begin{tabular}{lccc|ll|ll}
    \toprule
    ID & PS & DV & DB & \multicolumn{2}{c|}{CMU0} & \multicolumn{2}{c}{CMU3} \\
    & & & & AP25 $\uparrow$ & mAP $\uparrow$ & AP25 $\uparrow$ & mAP $\uparrow$ \\
    \midrule
    (a) & \xmark & \xmark & \xmark & 92.3 & 97.9 & 52.5 & 78.8 \\
    (b) & \cmark & \xmark & \xmark & 92.0 & 98.0 & 44.1 & 76.5 \\
    (c) & \cmark & \cmark & \xmark & 92.4 & 97.8 & 46.0 & 82.7 \\
    (d) & \xmark & \cmark & \cmark & 83.8 & 96.2 & 46.4 & 85.1 \\
    (e) & \cmark & \xmark & \cmark & 91.6 & 98.1 & 53.5 & 88.1 \\
    (f) & \cmark & \cmark & \cmark & $\textbf{93.7}$ & $\textbf{98.6}$ & $\textbf{64.3}$ & $\textbf{92.3} $  \\
    \bottomrule
  \end{tabular}
  \end{adjustbox}
\end{table}

Table~\ref{tab:ablation-depth} evaluates
the depth-prediction components:
depth embeddings (DE), pose-context modeling (PCM),
and target-depth masking (TDM).
Introducing DE (row~(b)) improves AP25/mAP by
3.0/0.7 points on CMU0 and 9.4/0.3 points on CMU3,
demonstrating the benefit of incorporating
metric reference-depth information.

Adding PCM without TDM (row~(c)) further improves
AP25 by 0.2 points on CMU0, while mAP remains unchanged.
On CMU3, however, AP25 decreases by 6.3 points,
despite a 1.1-point improvement in mAP.
Introducing TDM alongside PCM (row~(d))
increases AP25/mAP by 0.2/0.0 points on CMU0
and 13.6/2.8 points on CMU3 relative to row~(c).
Compared with DE alone (row~(b)),
the complete depth branch improves AP25/mAP
by 0.4/0.0 points on CMU0 and 7.3/3.9 points
on CMU3.
These results support the joint use of PCM
and TDM, particularly under challenging
camera configurations.

Further depth-error analysis in
Sec.~\ref{sec:suppl_depth_analysis} of the supplementary
material shows that introducing TDM reduces
the mean depth-prediction error on CMU3
from 62.0 to 38.3~mm for joints whose
ground truth is visible in exactly one camera.
This supports the role of TDM in improving
depth prediction for joints with limited
visibility.

\begin{table}[t]
  \centering
    \caption{Ablation of the depth-branch components:
    depth embedding (DE), target-depth masking (TDM),
    and pose-context modeling (PCM).}
  \label{tab:ablation-depth}
  \small
  \setlength{\tabcolsep}{3pt}
  \begin{adjustbox}{max width=\columnwidth}
  \begin{tabular}{lccc|ll|ll}
    \toprule
    ID & DE & PCM & TDM & \multicolumn{2}{c|}{CMU0} & \multicolumn{2}{c}{CMU3} \\
    & & & & AP25 $\uparrow$ & mAP $\uparrow$ & AP25 $\uparrow$ & mAP $\uparrow$ \\
    \midrule
    (a) & \xmark & \xmark & \xmark & 90.3 & 97.9 & 47.6 & 88.1 \\
    (b) & \cmark & \xmark & \xmark & 93.3 & \textbf{98.6} & 57.0 & 88.4 \\
    (c) & \cmark & \cmark & \xmark & 93.5 & \textbf{98.6} & 50.7 & 89.5 \\
    (d) & \cmark & \cmark & \cmark & $\textbf{93.7}$ & $\textbf{98.6}$ & $\textbf{64.3}$ & $\textbf{92.3} $ \\
    \bottomrule
  \end{tabular}
  \end{adjustbox}
\end{table}

\section{Conclusion and Future Work}
We presented BoE, a multi-view 3D human pose estimator
that augments geometric reconstruction with learned
monocular depth cues.
BoE reduces monocular depth ambiguity by conditioning
depth predictions on previously reconstructed 3D body
geometry.
LDAT incorporates predicted camera-frame depths as
additional constraints in differentiable algebraic
triangulation, enabling reconstruction even when only
a single camera contributes an observation.
Through iterative refinement, predicted depths and
reconstructed body geometry inform and improve each
other.
Our experiments demonstrate the benefits of conditioning
monocular depth prediction on reconstructed body geometry
and of incorporating predicted depths into triangulation
when multi-view evidence is limited.
On CMU Panoptic, BoE achieves state-of-the-art in-domain
performance and superior average performance across
evaluated camera subsets and unseen camera arrangements,
with a 10.6 AP25-point improvement over the previous
state of the art in the three-camera setting.

Nevertheless, the weaker transfer to Campus highlights
an opportunity to improve the cross-scene generalization
of depth prediction.
Moreover, while predicted depths enable reconstruction
from a single-view observation, BoE cannot recover
joints that are occluded in all cameras.
Future work could address this limitation by
incorporating kinematic models or learned 3D pose
priors to enable inference of completely unobserved
joints.

\textbf{Acknowledgements} This work was supported by AB Volvo, the Wallenberg AI, Autonomous Systems and Software Program (WASP) funded by the Knut and Alice Wallenberg Foundation, and the Vinnova funded projects SMILE IV (2023-00789), AIHURO (2022-03012) and ARTWORK (ITEA 22019).

\section*{LLM Usage Statement}
OpenAI's ChatGPT was used to assist with language editing and improving the clarity and presentation of the manuscript. OpenAI's Codex was used to assist with software implementation, including code generation and visualization. All AI-generated outputs were reviewed and validated by the authors, who take full responsibility for the research and its results.
{
    \small
    \bibliographystyle{ieeenat_fullname}
    \bibliography{main}
}

\clearpage
\setcounter{page}{1}
\maketitlesupplementary

\section{Implementation details}
\label{sec:suppl_arch}

Input images are resized to $960\times512$ and processed
by a ResNet-50 backbone pretrained for 2D pose estimation
on COCO.
Following~\cite{mvgformer}, we use 256-dimensional joint
tokens and initialize 1024 person queries on a regular
3D grid over the capture volume.
Queries with confidence below $\epsilon=0.1$ are filtered
between decoder layers, followed by non-maximum suppression
to remove redundant predictions. Training takes approximately 60 hours on a single node
equipped with four NVIDIA GH200 GPUs.

\paragraph{Training objective.}
We supervise the outputs of all $L=4$ decoder layers.
For regression, we follow MVGFormer~\cite{mvgformer}
and assign the $K=5$ nearest initial query anchors
to each ground-truth pose, keeping these assignments
fixed across decoder layers.
For classification, each predicted pose receives a
positive label if its MPJPE to the closest
ground-truth pose is below 150\,mm, and a negative
label otherwise.
Classification labels are recomputed independently
at each decoder layer.

The total training objective is
\begin{equation}
\mathcal{L}_{\mathrm{total}}
=
\sum_{\ell=1}^{L}
\left(
2\mathcal{L}_{\mathrm{cls}}^{(\ell)}
+5\mathcal{L}_{\mathrm{3D}}^{(\ell)}
+5\mathcal{L}_{\mathrm{2D}}^{(\ell)}
+5\mathcal{L}_{\mathrm{depth}}^{(\ell)}
\right).
\end{equation}
Here, $\mathcal{L}_{\mathrm{cls}}$ is a sigmoid focal
loss on the pose scores, with focal-loss parameters
$\alpha=0.25$ and $\gamma=2$.
$\mathcal{L}_{\mathrm{3D}}$ is a per-joint $\ell_1$
loss on reconstructed 3D poses in absolute
coordinates.
$\mathcal{L}_{\mathrm{2D}}$ and
$\mathcal{L}_{\mathrm{depth}}$ are masked per-view
$\ell_1$ losses on predicted 2D joint locations
and absolute camera-frame joint depths, respectively.
For both per-view losses, we project ground-truth
3D joints into each camera and exclude joints
whose projections fall outside the image bounds.

The regression losses are summed over matched queries
and normalized by the number of ground-truth persons.
The 3D loss is averaged over joint coordinates,
while the 2D and depth losses are additionally
averaged over the configured number of camera views.
Masked joints and dropped views contribute zero
error without changing these normalization factors.
The classification focal loss is summed over all
queries and normalized by
the number of ground-truth persons.

The 2D confidence scores and depth-confidence gates
receive no direct supervision; they are learned
through their contribution to the reconstructed
3D poses.

\paragraph{Query representation and initialization.}
The decoder maintains $N$ learnable pose queries, each
representing a candidate person through a set of $J$
3D joint locations and corresponding appearance tokens.
For query $n$, we denote the geometric representation by
$\mathbf{X}_n=\{\mathbf{x}_{n,j}\}_{j=1}^{J}$,
where $\mathbf{x}_{n,j}\in\mathbb{R}^{3}$ is the
world-coordinate position of joint $j$.
The appearance representation is
$\mathbf{F}_n=\{\mathbf{f}_{n,j}\}_{j=1}^{J}$,
where $\mathbf{f}_{n,j}\in\mathbb{R}^{D}$ is a
$D$-dimensional joint feature.

Following prior work~\cite{mvp,mvgformer,mvssm},
we define two sets of learnable embeddings:
$\mathcal{E}^{\mathrm{person}}
=\{\mathbf{e}^{\mathrm{person}}_n\}_{n=1}^{N}$
and
$\mathcal{E}^{\mathrm{joint}}
=\{\mathbf{e}^{\mathrm{joint}}_j\}_{j=1}^{J}$,
where all embeddings belong to $\mathbb{R}^{D}$.
The initial appearance token for joint $j$ of
query $n$ is constructed as
\begin{equation}
\mathbf{f}^{(0)}_{n,j}
=
\mathbf{e}^{\mathrm{person}}_n
+
\mathbf{e}^{\mathrm{joint}}_j.
\end{equation}
The person embedding is shared across all joints
of a query, while the joint embedding is shared
across all queries.

The initial geometric representation is constructed
from a fixed 3D T-pose template.
Let $\{\mathbf{t}_j\}_{j=1}^{J}$ denote the template
joint coordinates relative to its root, and let
$\mathbf{r}_n\in\mathbb{R}^{3}$ denote an initial
root location sampled from a regular 2D grid in the horizontal scene plane with fixed height.
The initial joint locations are
\begin{equation}
\mathbf{x}^{(0)}_{n,j}
=
\mathbf{r}_n+\mathbf{t}_j.
\end{equation}
Thus, the initial queries share the same skeletal
configuration but are distributed across different
candidate root locations in the scene.

At each decoder layer, the current 3D joint
locations serve as geometric references for
projective attention and depth prediction.
The decoder subsequently updates the geometric
representation through depth-aware triangulation
and refines the appearance tokens using the
extracted image features.

\paragraph{Geometry-conditioned depth prediction.}

\begin{table}[t]
  \centering
\caption{\textbf{Depth prediction error across decoder layers.}
Mean per-camera depth error (mm) on CMU0 and CMU3.
All denotes all evaluated joints, while Single-view
denotes joints whose ground truth is visible in exactly
one camera. A dash indicates that no joints satisfy
the corresponding criterion.}
  \label{tab:paper-results-11}
  \small
  \setlength{\tabcolsep}{3pt}
  \begin{adjustbox}{max width=\textwidth}
\begin{tabular}{l|cc|cc}
  \toprule
  & \multicolumn{2}{c|}{CMU0}
  & \multicolumn{2}{c}{CMU3} \\
  Decoder layer
  & All & Single-view
  & All & Single-view \\
  \midrule
  0 & 77.4 & -- & 98.7 & 55.6 \\
  1 & 42.1 & -- & 43.9 & 55.7 \\
  2 & 30.4 & -- & 35.0 & 45.0 \\
  3 & 27.5 & -- & 34.1 & 38.3 \\
  \bottomrule
\end{tabular}
  \end{adjustbox}
\end{table}

\begin{table}[t]
  \centering
  \caption{\textbf{Depth-branch component ablation.}
Final-layer mean per-camera depth error (mm) on CMU0
and CMU3 for different combinations of depth embedding
(DE), pose-context modeling (PCM), and target-depth
masking (TDM). All denotes all evaluated joints, while
Single-view denotes joints whose ground truth is visible
in exactly one camera. A dash indicates that no joints
satisfy the corresponding criterion.}
  \label{tab:paper-results-depth-ablation-l3}
  \small
  \setlength{\tabcolsep}{3pt}
  \begin{adjustbox}{max width=\textwidth}
\begin{tabular}{lccc|cc|cc}
  \toprule
  & & & & \multicolumn{2}{c|}{CMU0}
  & \multicolumn{2}{c}{CMU3} \\
  ID & DE & PCM & TDM
  & All & Single-view
  & All & Single-view \\
  \midrule
  (a) & \xmark & \xmark & \xmark
      & 73.5 & -- & 145.4 & 77.0 \\
  (b) & \cmark & \xmark & \xmark
      & 24.8 & -- & 35.8 & 74.5 \\
  (c) & \cmark & \cmark & \xmark
      & 29.3 & -- & 34.6 & 62.0 \\
  (d) & \cmark & \cmark & \cmark
      & 27.5 & -- & 34.1 & 38.3 \\
  \bottomrule
\end{tabular}
  \end{adjustbox}
\end{table}

\emph{Depth embedding.}
The depth-embedding network $e_z(\cdot )$ consists of two linear layers with a ReLU activation in between. The hidden dimension is the same as the output dimension, i.e., 256. 

\emph{Pose-context modeling.}
For each camera $v$, person query $n$, and target joint $j$, the encoder
$\operatorname{E}(\cdot)$ (in Eq. \ref{eq:depth-contextualized-feature}) processes the masked joint sequence
$\mathbf{s}^{(j)}_{v,n}=\{\mathbf{s}^{(j)}_{v,n,k}\}_{k=1}^{J}$ using
three stacked Transformer-style blocks. Let
$\mathbf{s}^{(j,0)}_{v,n}=\mathbf{s}^{(j)}_{v,n}$. In block $\ell$, we
first apply joint-wise multi-head self-attention, followed by a residual
connection and layer normalization:
\begin{equation}
\mathbf{H}^{(j,\ell)}_{v,n}
=
\operatorname{LN}\!\left(
\mathbf{s}^{(j,\ell-1)}_{v,n}
+ \operatorname{Dropout}\!\left(
\operatorname{MHSA}\!\left(\mathbf{s}^{(j,\ell-1)}_{v,n}\right)
\right)
\right).
\end{equation}
We then apply a position-wise feed-forward network, again with a residual
connection and layer normalization:
\begin{equation}
\mathbf{s}^{(j,\ell)}_{v,n}
=
\operatorname{LN}\!\left(
\mathbf{H}^{(j,\ell)}_{v,n}
+ \operatorname{FFN}\!\left(\mathbf{H}^{(j,\ell)}_{v,n}\right)
\right),
\qquad \ell=1,2,3.
\label{eq:PCM2}
\end{equation}
The FFN consists of a linear projection from $D$ to $D_{\mathrm{ffn}}$, a
ReLU activation, dropout, a linear projection from $D_{\mathrm{ffn}}$ to
$D$, and a final dropout. In
our implementation, $D=256$ and $D_{\mathrm{ffn}}=1024$. The
output of the encoder is
$\operatorname{E}(\mathbf{s}^{(j)}_{v,n})=
\mathbf{s}^{(j,3)}_{v,n}$, from which we retain the feature at the target
joint index $j$. Thus, self-attention operates only among the $J$ joints of
one person in one camera view; it does not mix features across persons or
views.

\emph{Depth and confidence prediction heads.}
The target-joint feature output by the pose-context encoder,
$\mathbf{h}_{v,n,j}=[\operatorname{E}(\mathbf{s}^{(j)}_{v,n})]_j$, is
fed to two independent three-layer MLPs: a depth head $f_{\mathrm{D}}$ and
a depth-confidence head $f_{\mathrm{gate}}$. Each MLP has the architecture
$256 \rightarrow 256 \rightarrow 256 \rightarrow 1$, with ReLU activations
after its first two linear layers and a linear scalar output. 
Unlike the PCM encoder, the depth and confidence MLPs do not use dropout

\paragraph{Target feature update.}
For completeness, we detail the implementation of
$\operatorname{FFN}(\mathbf{f}_{n,j},\bar{\mathbf{a}}_{n,j})$ in
Eq.~\eqref{eq:query-feature-update}. The mean-pooled projective-attention
feature is first projected and added to the appearance token through a
residual update:
\begin{equation}
\mathbf{r}_{n,j}
=
\operatorname{LN}\!\left(
\mathbf{f}_{n,j}
+ \operatorname{Dropout}\!\left(
\operatorname{Linear}_{D\rightarrow D}
\left(\bar{\mathbf{a}}_{n,j}\right)
\right)
\right).
\end{equation}
This is followed by a standard residual feed-forward update:
\begin{equation}
\mathbf{f}'_{n,j}
=
\operatorname{LN}\!\left(
\mathbf{r}_{n,j}
+ \operatorname{FFN}\!\left(\mathbf{r}_{n,j}\right)
\right).
\end{equation}
This FFN has the same architecture as in the PCM (Eq. \ref{eq:PCM2}), but uses independent
parameters. 

\paragraph{Pose scoring implementation.}
For each person query $n$, we prepend a learned classification token
$\mathbf{c}\in\mathbb{R}^{D}$ to its updated joint appearance tokens:
\begin{equation}
\mathbf{Z}_{n}=
\left[\mathbf{c};\mathbf{f}_{n,1};\ldots;\mathbf{f}_{n,J}\right].
\end{equation}
We apply multi-head self-attention over the resulting $J+1$ tokens:
\begin{equation}
\mathbf{Z}'_{n}=
\operatorname{LN}\!\left(
\mathbf{Z}_{n}+
\operatorname{Dropout}\!\left(\operatorname{MHSA}(\mathbf{Z}_{n})\right)
\right).
\end{equation}
The updated classification token is mapped through a single linear layer to a pose confidence, which is used for filtering low-scoring queries after each decoder layer. This pose-scoring block uses $D=256$, eight
attention heads, and dropout probability $0.1$.

\section{Depth Prediction Analysis}
\label{sec:suppl_depth_analysis}

\paragraph{Depth prediction across decoder layers.}
Table~\ref{tab:paper-results-11} reports the mean per-camera
depth-prediction error at each decoder layer on CMU0 and CMU3,
both across all evaluated joints and for joints whose ground
truth is visible in exactly one camera.
The all-joint error decreases across decoder layers,
from 77.4 to 27.5\,mm on CMU0 and from 98.7 to 34.1\,mm
on CMU3.
For joints whose ground truth is visible in exactly one
camera on CMU3, the error also decreases substantially,
reaching 38.3\,mm at the final layer.
These results indicate that iterative pose refinement
progressively improves depth estimation, including for
joints with limited ground-truth visibility.

\paragraph{Depth-prediction component ablation.}
Table~\ref{tab:paper-results-depth-ablation-l3} examines the
effects of depth embeddings (DE), pose-context modeling (PCM),
and target-depth masking (TDM) on final-layer depth-prediction
error.
Adding DE reduces the all-joint error from 73.5 to 24.8\,mm
on CMU0 and from 145.4 to 35.8\,mm on CMU3.
Adding PCM without TDM increases the error on CMU0 to
29.3\,mm, while slightly reducing it on CMU3 to 34.6\,mm.
Introducing TDM improves the all-joint errors to 27.5\,mm
on CMU0 and 34.1\,mm on CMU3, while substantially reducing
the error from 62.0 to 38.3\,mm on CMU3 for joints whose
ground truth is visible in exactly one camera.
These results highlight the importance of masking the
target joint's reference depth when leveraging geometric
context, particularly for joints with limited multi-view
observations.

\paragraph{Cross-scene depth prediction.}
Table~\ref{tab:cross-scene-depth-error} reports final-layer
depth-prediction errors on Shelf and Campus.
The mean error is 61.7\,mm on Shelf, compared with
27.5\,mm on CMU0 and 34.1\,mm on CMU3, but rises to
6,583.4\,mm on Campus.
This discrepancy coincides with a substantial shift in
subject depths: typical ground-truth depths are approximately
2--5\,m on CMU Panoptic and Shelf, compared with
5--15\,m on Campus.
The results indicate that the learned depth predictor
generalizes poorly to the substantially different depth
distribution on Campus.

\begin{table}[t]
  \centering
  \caption{\textbf{Cross-scene depth prediction error.}
  Final-layer mean per-camera depth error on Campus
  and Shelf without fine-tuning. Neither dataset
contains evaluated joints whose ground truth is
visible in exactly one camera.}
  \label{tab:cross-scene-depth-error}
  \small
  \setlength{\tabcolsep}{5pt}
  \begin{tabular}{lc}
    \toprule
    Dataset & Mean depth error (mm) \\
    \midrule
    Campus & 6,583.4 \\
    Shelf  & 61.7 \\
    \bottomrule
  \end{tabular}
\end{table}

\section{Depth-aware triangulation error}
\label{sec:suppl_da_triangulation}

\begin{figure*}
    \centering
    \includegraphics[width=0.8\linewidth]{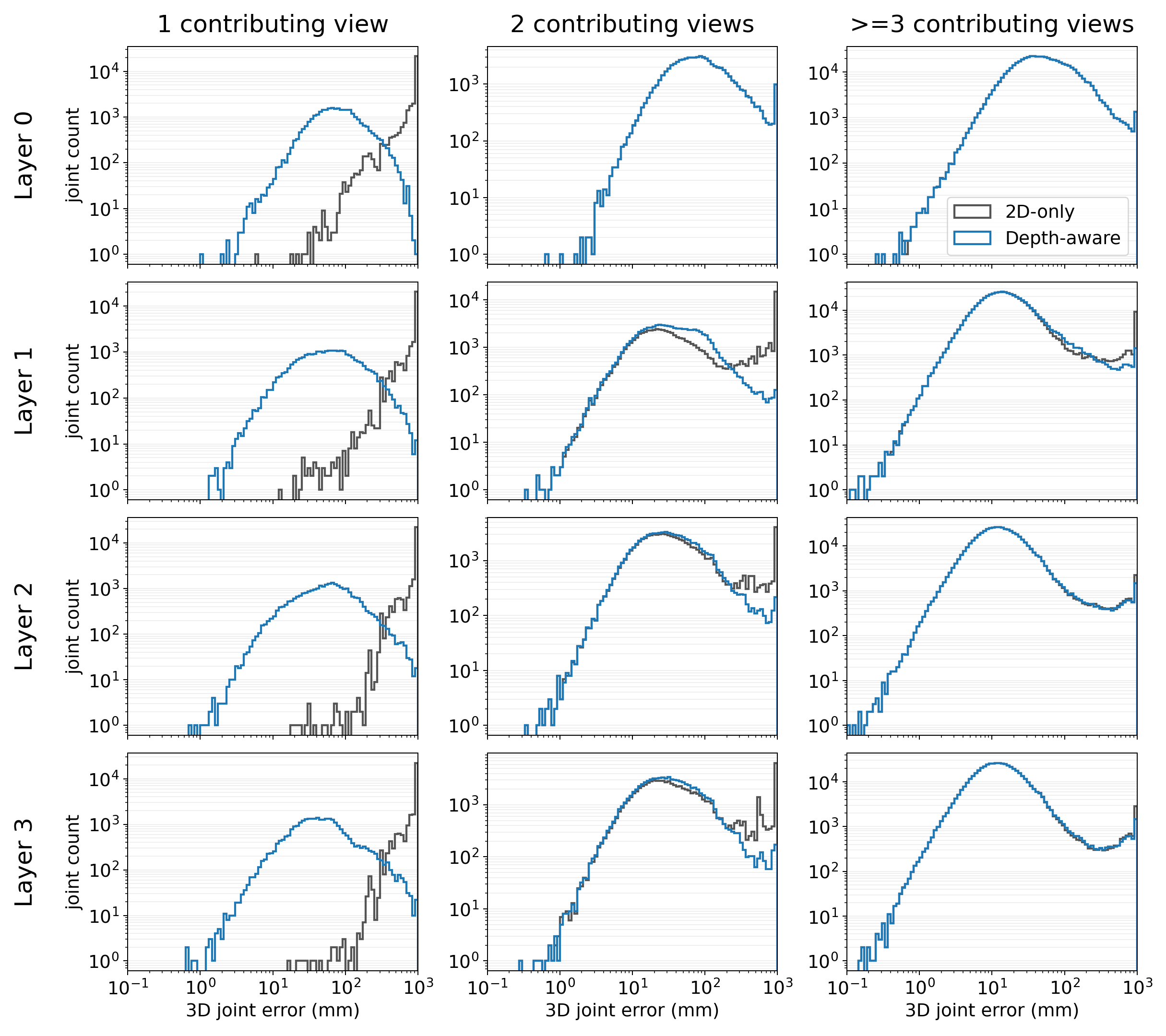}
\caption{\textbf{Effect of depth constraints on triangulation.}
Distributions of 3D joint-position errors on CMU3 using
LDAT (blue) and conventional
2D-only DLT (gray). The latter is obtained by removing the depth
constraints from LDAT while keeping the predicted 2D
observations fixed. Rows correspond to decoder layers,
and columns group joints by the number of contributing
cameras, defined by the views used for triangulation
rather than ground-truth visibility.}
    \label{fig:decoder_layer_error}
\end{figure*}

Figure~\ref{fig:decoder_layer_error} compares the distributions of 3D joint
errors obtained using depth-aware triangulation and conventional
2D-only DLT on the same CMU3 predictions.
Results are grouped by decoder layer (rows) and number of contributing
cameras (columns): one, two, or three to four cameras.
For this comparison, the depth constraints are removed from the
triangulation system while keeping the predicted 2D observations fixed.

The largest improvements from depth-aware triangulation
occur when only one camera contributes to reconstruction.
The benefit remains evident with two cameras, while
LDAT and conventional triangulation yield increasingly
similar results with three or four contributing cameras.
This is expected, as additional 2D observations provide
stronger geometric constraints, reducing the reliance
on monocular depth estimates.
These observations support the value of monocular depth
constraints when multi-view observations are limited.

\section{Additional qualitative results}
\label{sec:suppl_qualitative}
\begin{figure*}[p]
    \centering
    \qualrow{\hspace{15pt} Ours \hspace{32pt} MVGFormer}{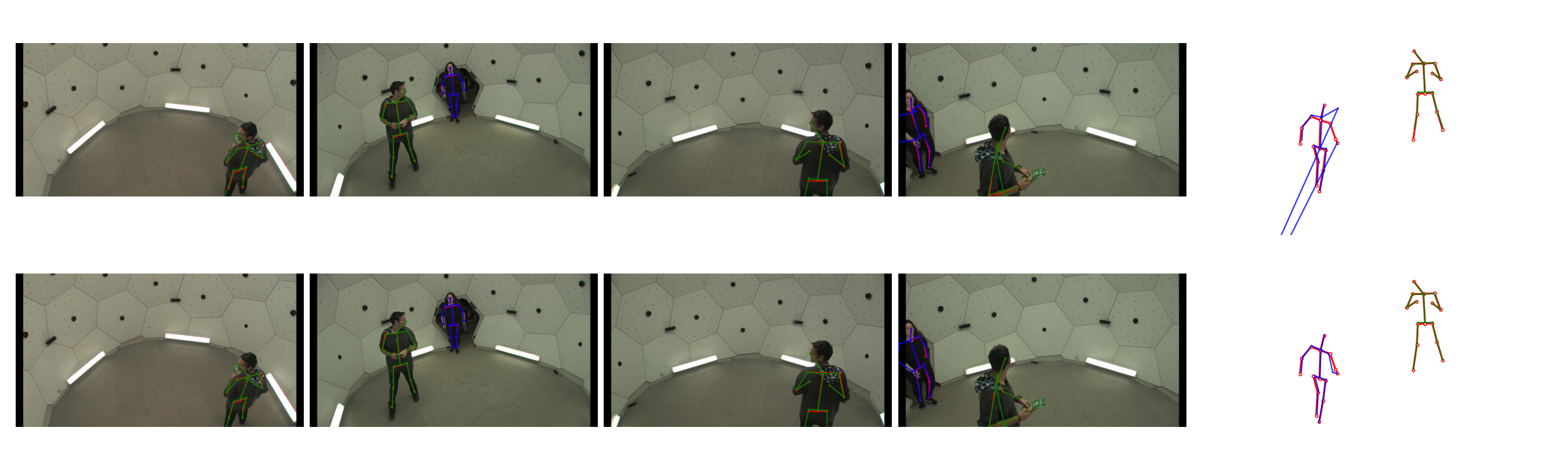}
    \qualrow{\hspace{15pt} Ours \hspace{32pt} MVGFormer}{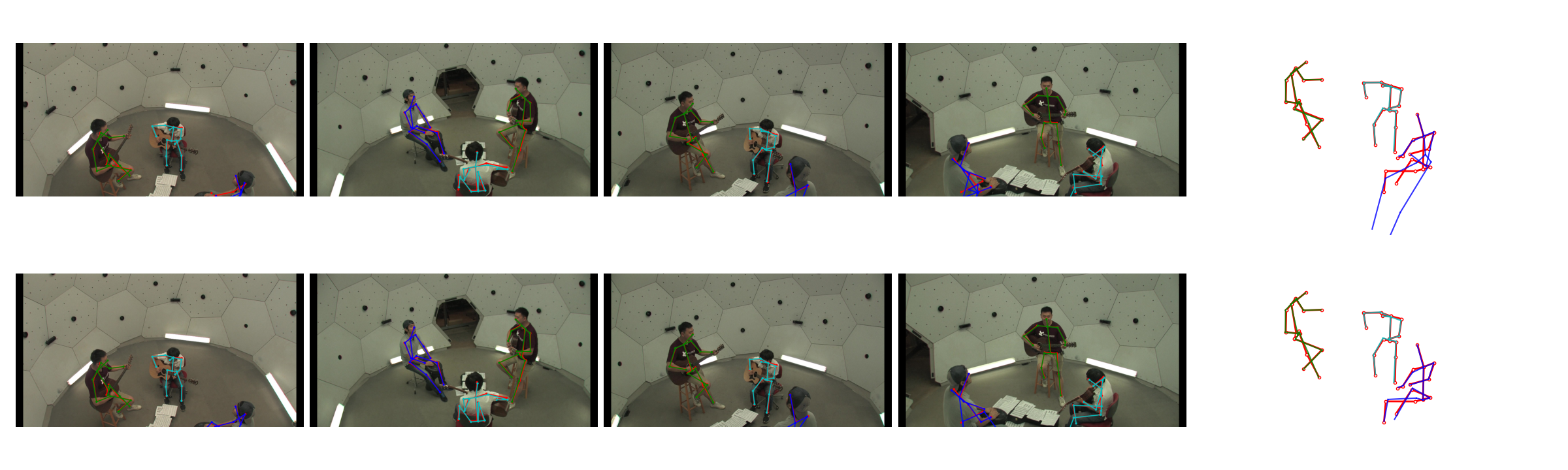}
    \qualrow{\hspace{15pt} Ours \hspace{32pt} MVGFormer}{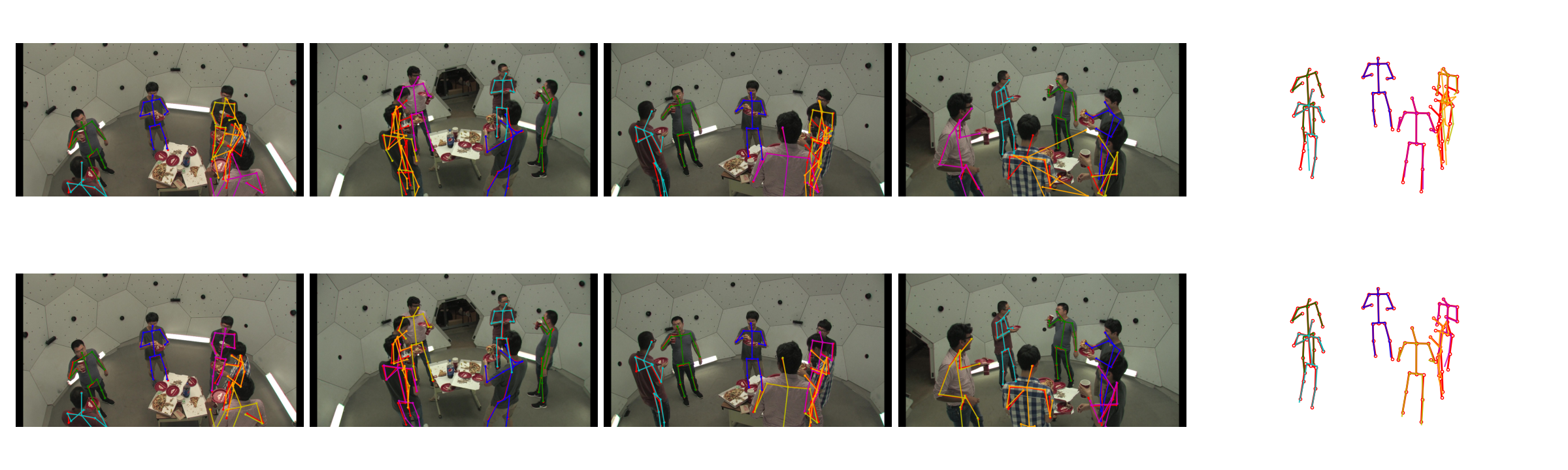}
    \qualrow{\hspace{15pt} Ours \hspace{32pt} MVGFormer}{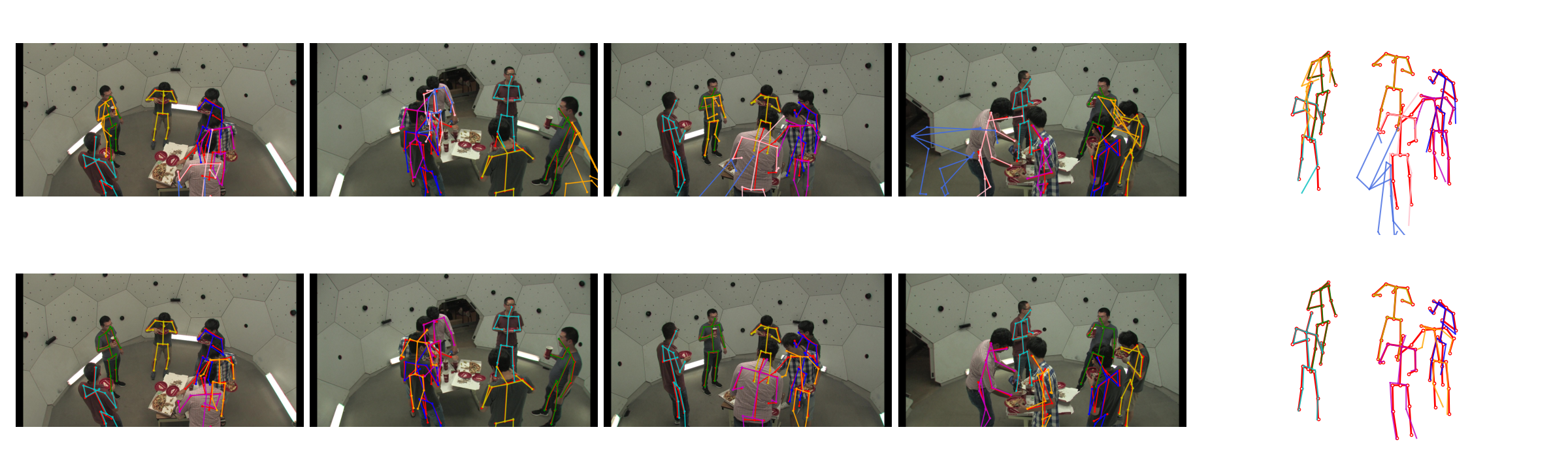}
\caption{\textbf{Additional qualitative comparisons
of BoE and MVGFormer on CMU3.}
RGB images with predicted 2D poses and projected
ground-truth poses are shown on the left, and
reconstructed 3D poses on the right.
Ground truth is shown in red, while predictions
are shown in other colors.
The examples correspond to the first four 3D pose
comparisons in Fig.~\ref{fig:qual-3d} of the main paper.}
    \label{fig:qual_failures0}
\end{figure*}

\begin{figure*}[p]
    \centering
    \qualrow{\hspace{15pt} Ours \hspace{32pt} MVGFormer}{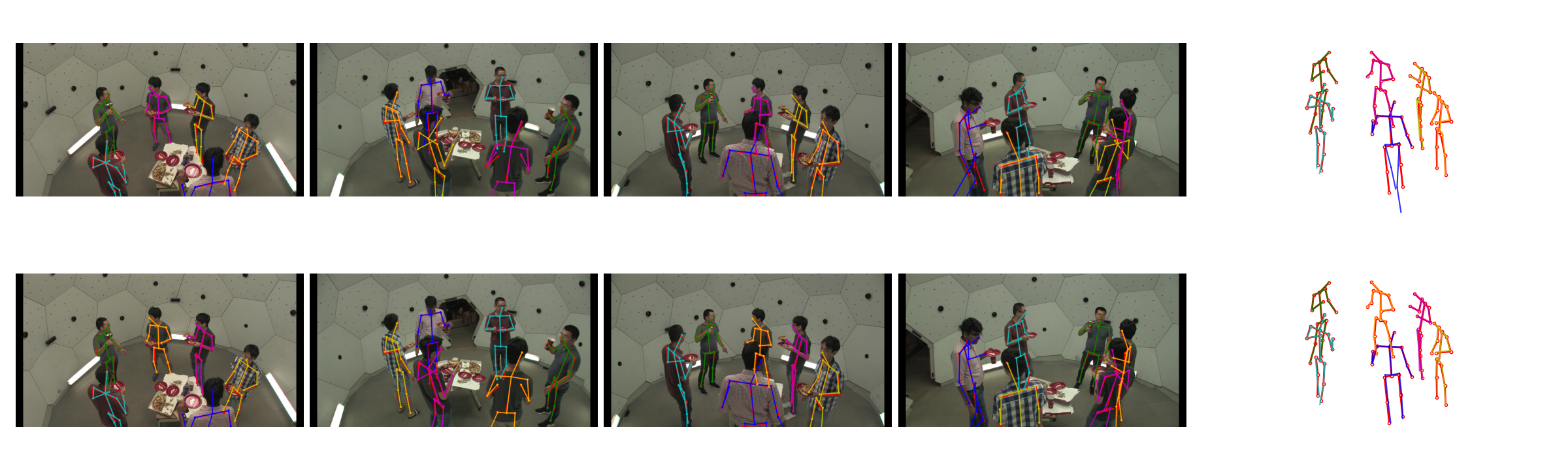}
    \qualrow{\hspace{15pt} Ours \hspace{32pt} MVGFormer}{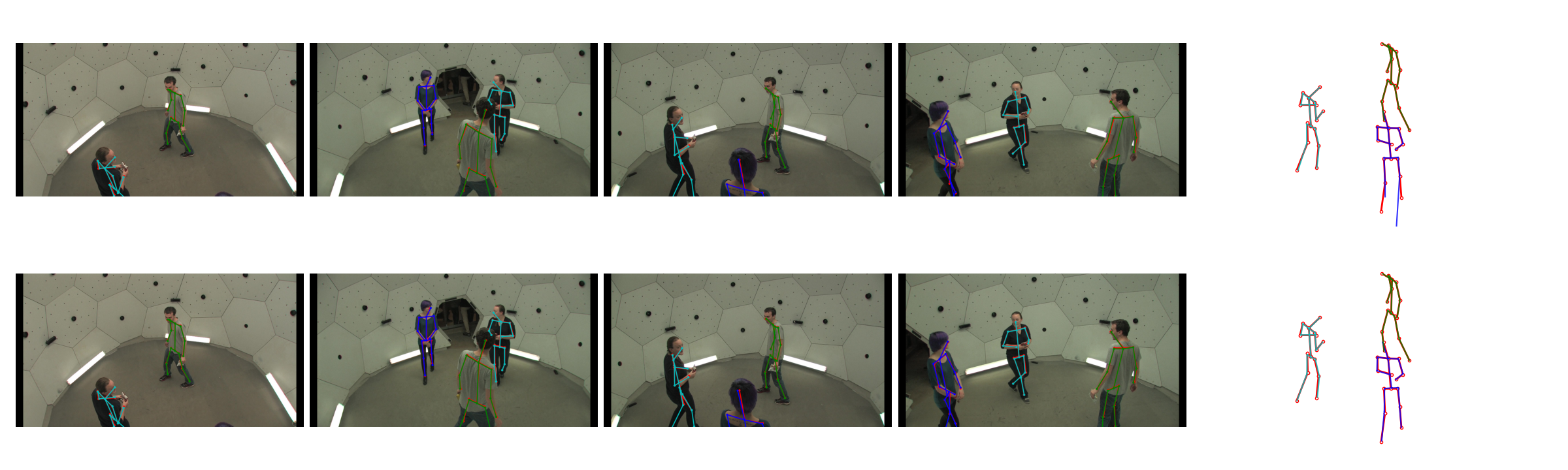}
    \qualrow{\hspace{15pt} Ours \hspace{32pt} MVGFormer}{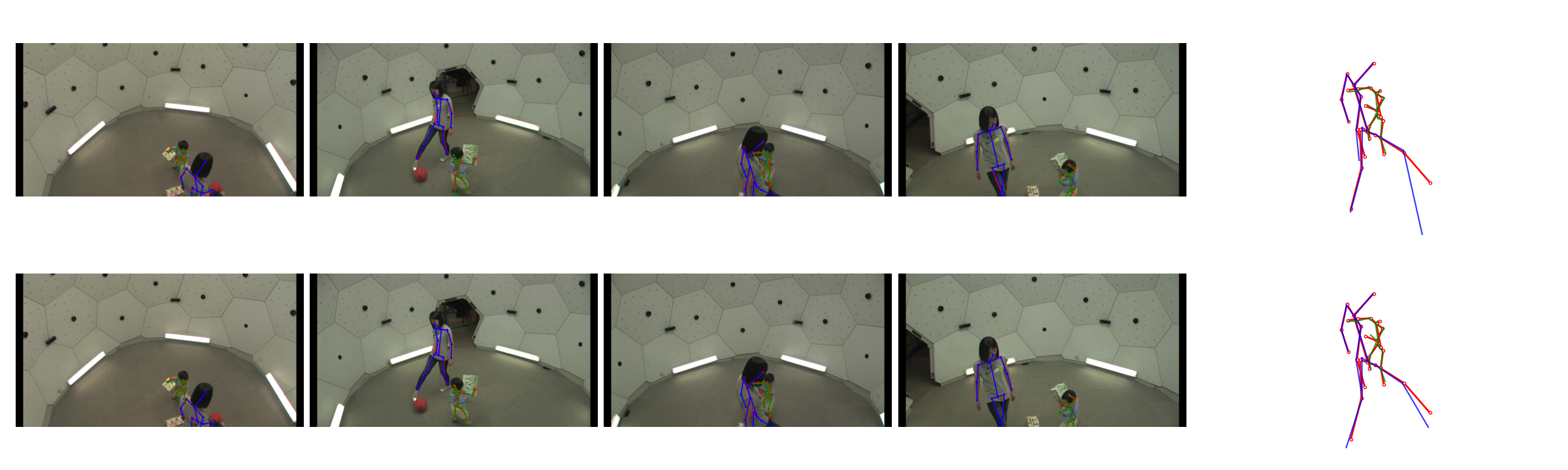}
    \qualrow{\hspace{15pt} Ours \hspace{32pt} MVGFormer}{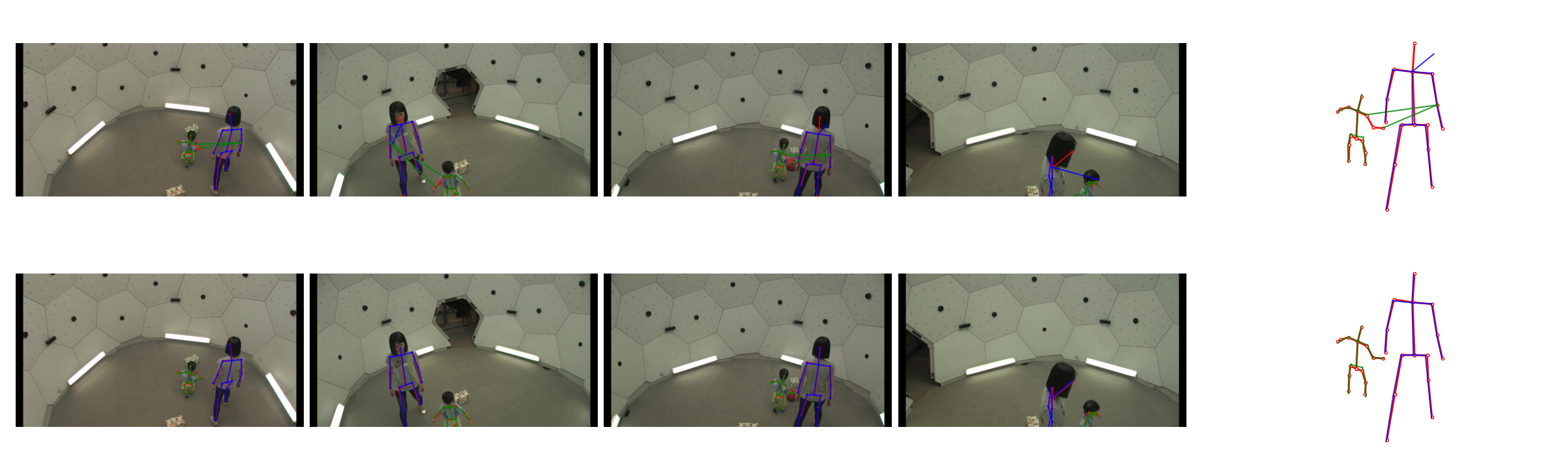}
\caption{\textbf{Additional qualitative comparisons
of BoE and MVGFormer on CMU3 (continued).}
RGB images with predicted 2D poses and projected
ground-truth poses are shown on the left, and
reconstructed 3D poses on the right.
Ground truth is shown in red, while predictions
are shown in other colors.
The examples correspond to the next four 3D pose
comparisons in Fig.~\ref{fig:qual-3d} of the main paper.}
    \label{fig:qual_failures1}
\end{figure*}

\begin{figure*}[p]
    \centering
    \qualrow{\hspace{15pt} Ours \hspace{32pt} MVGFormer}{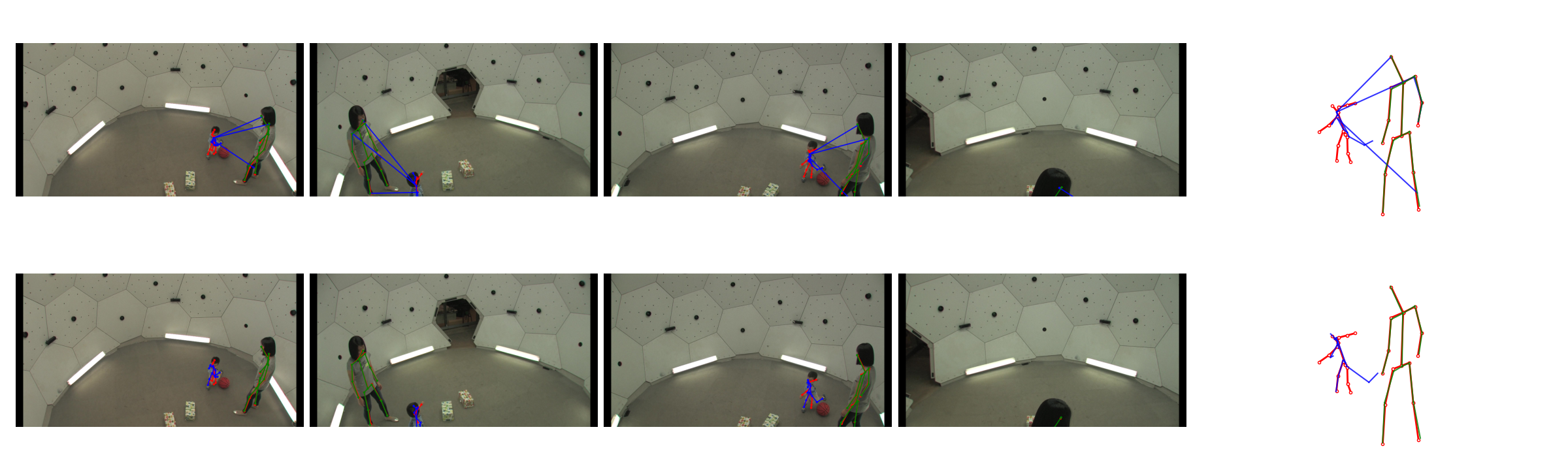}
    \qualrow{\hspace{15pt} Ours \hspace{32pt} MVGFormer}{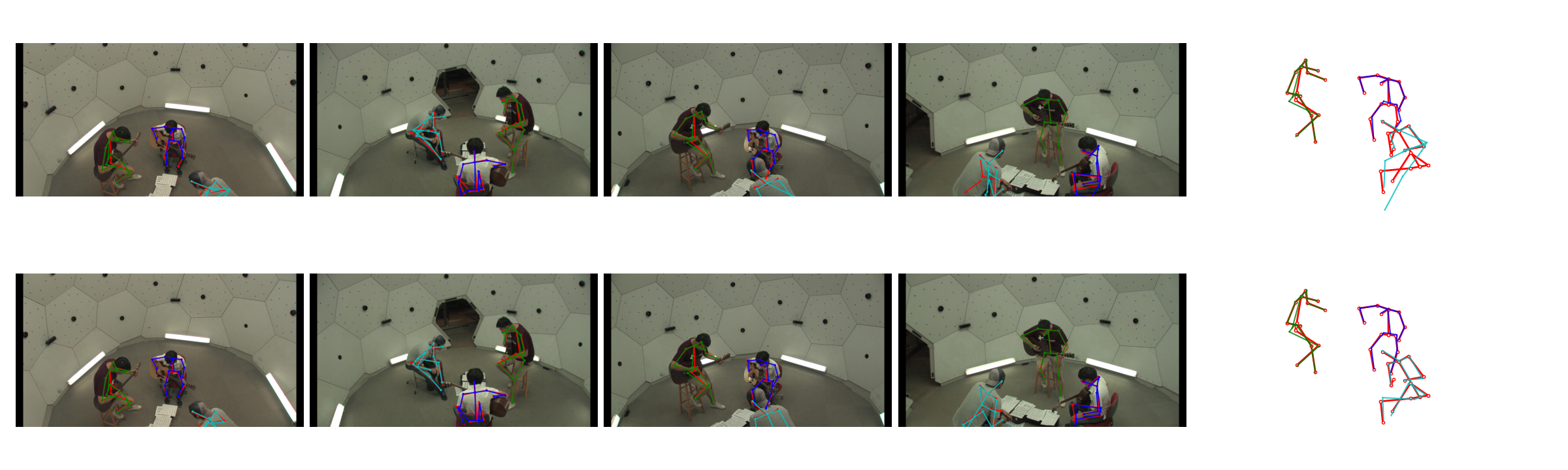}
\caption{\textbf{Additional qualitative comparisons
of BoE and MVGFormer on CMU3 (continued).}
RGB images with predicted 2D poses and projected
ground-truth poses are shown on the left, and
reconstructed 3D poses on the right.
Ground truth is shown in red, while predictions
are shown in other colors.
The examples correspond to the final two 3D pose
comparisons in Fig.~\ref{fig:qual-3d} of the main paper.}
    \label{fig:qual_failures2}
\end{figure*}

\paragraph{CMU Panoptic.}
Figures~\ref{fig:qual_failures0}--\ref{fig:qual_failures2}
provide additional qualitative comparisons of BoE and
MVGFormer~\cite{mvgformer}, corresponding to the ten
3D pose comparisons shown in Fig.~\ref{fig:qual-3d}
of the main paper.

\begin{figure*}
    \centering
    \qualrow{\hspace{15pt} Ours \hspace{28pt} MVGFormer}{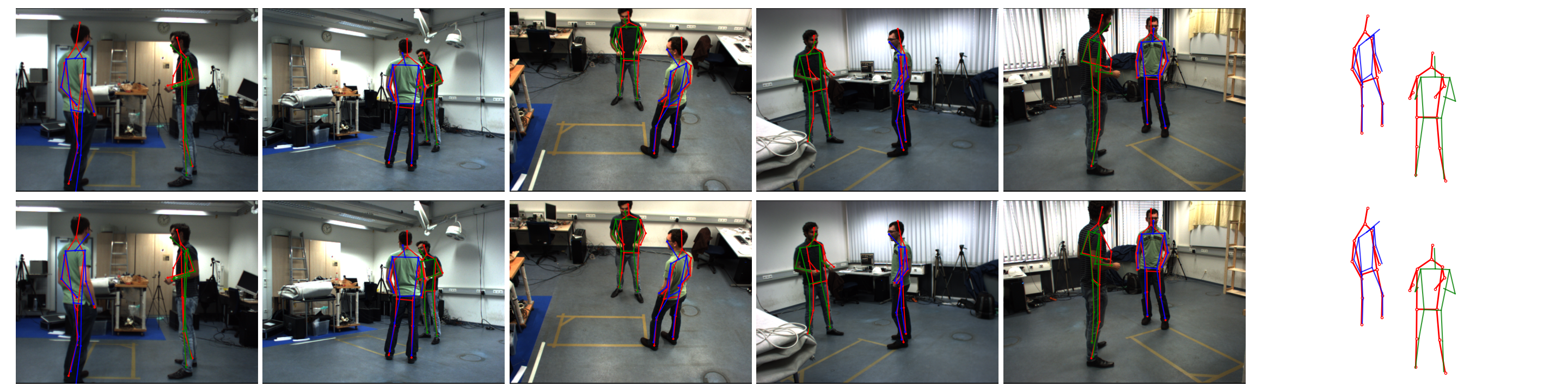}
\caption{\textbf{Qualitative comparison of BoE and
MVGFormer on Shelf.}
RGB images with predicted 2D poses and projected
ground-truth poses are shown on the left, and
reconstructed 3D poses on the right.
Ground truth is shown in red, while predictions
are shown in other colors.
Both methods produce similar 3D reconstructions
in this example.}
    \label{fig:shelf}
\end{figure*}

\begin{figure*}
    \centering
    \qualrow{\hspace{10pt} Ours \hspace{40pt} MVGFormer}{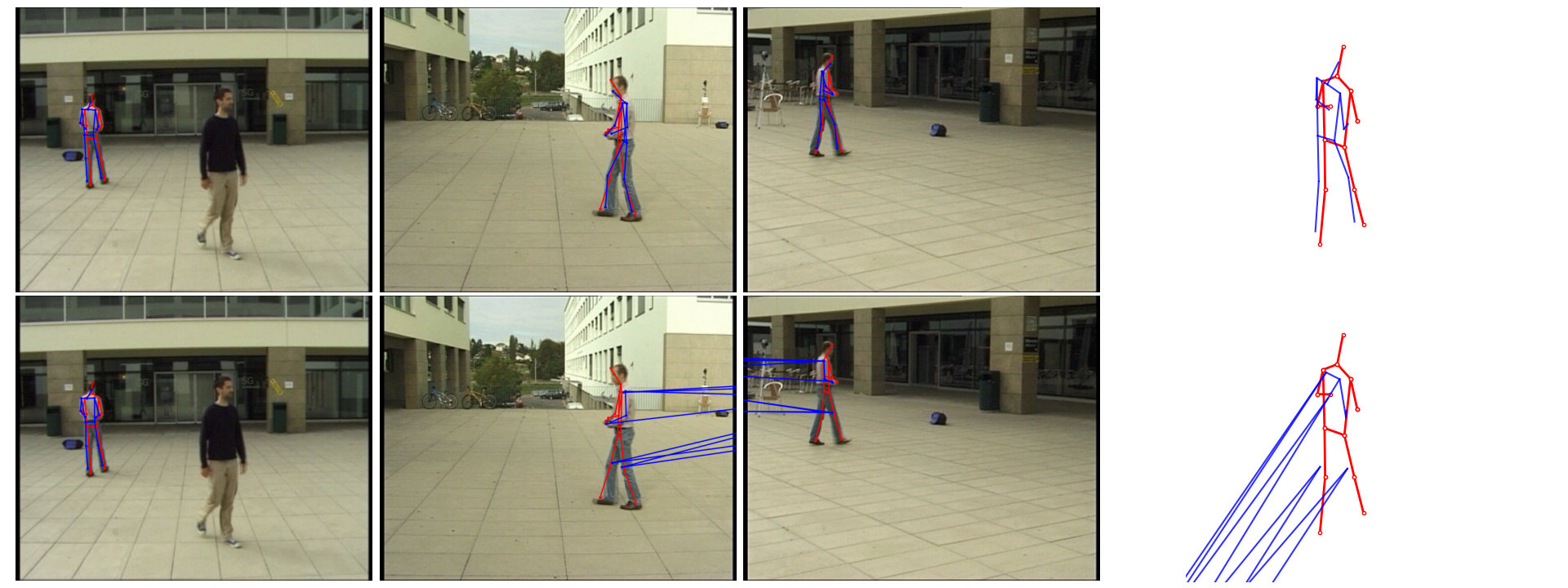}
\caption{\textbf{Qualitative comparison of BoE and
MVGFormer on Campus.}
RGB images with predicted 2D poses and projected
ground-truth poses are shown on the left, and
reconstructed 3D poses on the right.
Ground truth is shown in red, while predictions
are shown in blue.
BoE exhibits substantial reconstruction errors for
several joints whose depths are underestimated
in the leftmost camera.}
    \label{fig:campus}
\end{figure*}

\paragraph{Shelf and Campus}
\label{sec:suppl_shelfcampus}
Figures~\ref{fig:shelf} and~\ref{fig:campus}
compare DeTriPose and MVGFormer on Shelf and Campus,
respectively.
In the illustrated Shelf example, the two methods produce
similar reconstructions.
In the Campus example (Fig.~\ref{fig:campus}),
the predicted depths for several joints in the
leftmost camera are substantially smaller than
their ground-truth depths.
The reconstructed 3D positions of these joints
appear to be strongly influenced by the
underestimated depths, causing their projections
to fall outside the fields of view of the other
two cameras and preventing consistent multi-view
support.
However, other joints are correctly detected
across all three cameras, indicating that this
failure affects only part of the reconstructed pose.

\end{document}